\documentclass[11pt]{article}

\usepackage[a4paper,margin=1in]{geometry}
\usepackage[T1]{fontenc}
\usepackage{textcomp}
\usepackage[utf8]{inputenc}
\usepackage{mathptmx}
\usepackage{microtype}
\usepackage{setspace}
\usepackage{amsmath,amssymb,amsfonts}
\usepackage{graphicx}
\usepackage{booktabs}
\usepackage{array}
\usepackage[font=small,labelfont=bf]{caption}
\usepackage{enumitem}
\setlist{itemsep=2pt,topsep=4pt}
\usepackage{algorithm}
\usepackage{algpseudocode}
\usepackage{xcolor}
\usepackage[edges]{forest}

\usepackage{listings}
\usepackage{authblk}

\usepackage[numbers,sort&compress]{natbib}
\usepackage[hidelinks]{hyperref}
\usepackage{xurl}
\newcommand{\Description}[1]{}

\title{\textbf{Reimagining Text-to-SQL: A Survey of Large Language Model Architectures and Strategies}}

\author[1]{Shikhar Dave}
\author[1]{Bhaskarjit Sarmah}
\author[2]{Stefano Pasquali}
\affil[1]{Domyn, New Delhi, India}
\affil[2]{Domyn, New York, NY, United States}
\affil[ ]{\texttt{shikhar.dave@domyn.com}, \texttt{bhaskarjit.sarmah@domyn.com}, \texttt{stefano.pasquali@domyn.com}}
\date{}

\begin{document}

\maketitle

\begin{abstract}
\noindent Text-to-SQL systems translate natural language questions into executable SQL queries, democratizing access to relational data. Early approaches relied on handcrafted rules and template matching, while subsequent neural methods introduced encoder--decoder architectures, schema linking, and constrained decoding to handle increasingly complex queries. The emergence of large language models (LLMs), exemplified by systems such as ReFoRCE, Agentar-Scale-SQL, Snowflake's Arctic-Text2SQL-R1, and the modular XiYan-SQL framework, has dramatically advanced the field, pushing execution accuracy to new state-of-the-art levels on challenging benchmarks such as Spider, Spider~2.0, and BIRD. In parallel, modular frameworks decompose the task into sub-problems (schema retrieval, sketch-based generation, repair, and ranking), while agentic systems coordinate multiple specialized agents that communicate and dynamically allocate sub-tasks. This survey systematically reviews these LLM-driven approaches within a unified taxonomy of five paradigms (in-context learning, supervised and parameter-efficient fine-tuning, agentic and tool-augmented orchestration, retrieval-augmented generation, and reinforcement learning) and characterizes them by model granularity (monolithic vs.\ modular), level of supervision (zero-shot, fine-tuned, or reinforcement-based), and reasoning style (direct decoding vs.\ decomposed multi-step reasoning). We further analyze evaluation metrics, datasets, and open challenges, including domain generalization, efficiency, and real-world deployment, and we examine the role of the semantic layer in production systems. As LLMs increasingly access sensitive databases, we also discuss responsible AI practices. We survey over 100 papers and technical resources, focusing on work published between 2022 and 2026.

\end{abstract}

\section{Introduction}
\label{sec:intro}

Relational databases are the backbone of data infrastructure in industries such as finance, healthcare, logistics, and e-commerce, where vast volumes of structured data are queried daily to support decision-making and operations. But working with this data usually means writing SQL, which is a real barrier for business analysts, product managers, and other non-technical stakeholders who need answers from the data but do not necessarily have the training to query it themselves. According to recent industry statistics, around two in three people in the global workforce have access to business intelligence tools, but only 21\% are confident in their ability to work with data in impactful ways~\cite{howarth2023bi-stats}.
Text-to-SQL systems address this challenge by translating natural language questions into executable SQL, thereby democratizing data access. By offering conversational interfaces to complex datasets, these technologies can simplify analytics workflows, lower engineering costs, and accelerate insight generation across a wide range of enterprise applications.

Most early Text-to-SQL solutions were template- or rule-based, relying on fixed SQL structures and manually designed grammars. These systems worked for narrow domains, but they did not scale and struggled with complex or previously unseen schemas. Semantic parsing and schema understanding improved significantly with the emergence of deep learning, particularly encoder--decoder models based on LSTMs and Transformers. Models such as Seq2SQL~\cite{zhong2017seq2sql}, SQLNet~\cite{xu2017sqlnet}, and RAT-SQL~\cite{wang2019rat} leveraged structured decoding, pointer networks~\cite{vinyals2015pointer}, and schema-linking techniques to handle more realistic and compositional queries.

More recently, powerful large language models (LLMs) such as T5~\cite{raffel2019exploring}, Codex~\cite{chen2021evaluating}, PaLM~\cite{chowdhery2022palm}, and GPT-4~\cite{openai2023gpt4} have transformed the landscape of Text-to-SQL by enabling generalizable and highly accurate SQL generation across a wide variety of schemas and query styles. Modern LLM-based Text-to-SQL systems can be broadly grouped into three families:

\begin{itemize}
\item \textbf{In-context learning (ICL) approaches.} These use frozen LLMs with carefully designed prompts, often leveraging few-shot examples, SQL sketches, or demonstration chains. Advances such as \textbf{Chain-of-Thought (CoT)} prompting~\cite{wei2022chainofthought}, \textbf{self-consistency decoding}~\cite{wang2022selfconsistency}, and \textbf{SQL skeleton prompting}~\cite{guo2023prompting} have improved reasoning and compositional generalization without updating model weights.

\item \textbf{Fine-tuning approaches.} These methods adapt pre-trained language models to the Text-to-SQL task using supervised datasets such as Spider~\cite{yu2018spider} or BIRD~\cite{li2023llm}. Two strategies dominate: \textbf{full-parameter fine-tuning}, which updates all model weights to specialize the model for SQL generation, and \textbf{parameter-efficient fine-tuning (PEFT)}, which freezes most weights and trains lightweight modules (e.g., LoRA~\cite{hu2021lora} or adapters) to reduce compute and memory costs while retaining performance.

\item \textbf{Agentic and modular approaches.} Inspired by tool-augmented reasoning, these methods decompose Text-to-SQL into specialized sub-components such as schema linking, sub-query generation, correction, and ranking. Frameworks such as \textbf{XiYan-SQL}~\cite{gao2024xiyan} adopt a multi-generator ensemble with dedicated refiners and selectors, while agent-based methods such as \textbf{MAC-SQL}~\cite{wang2023macsql} and \textbf{SQLFixAgent}~\cite{cen2024sqlfixagent} integrate execution feedback and iterative self-correction. These systems often use external tools, database lookups, or validators to improve reliability.
\end{itemize}

\noindent A fourth family, \textbf{reinforcement learning (RL) with execution-based rewards}, has emerged since 2025 and now underpins many of the strongest systems on public leaderboards; we treat it as a distinct paradigm in this survey. Industry-grade models such as \textbf{Arctic-Text2SQL-R1}~\cite{yao2025arctic} from Snowflake illustrate the practical promise of these approaches, achieving state-of-the-art execution accuracy at the time of release. By combining large-scale pretraining with structured decoding, task decomposition, execution feedback, and careful prompt design, modern LLM-based Text-to-SQL systems achieve substantially better scalability and accuracy, reduce error rates on complex queries, and are becoming increasingly practical for enterprise settings.

\subsection{Relationship to Existing Surveys}

Several recent surveys have reviewed LLM-based Text-to-SQL methods from complementary perspectives.
Hong et al.~\cite{hong2025next} present a systematic review organized around question understanding, schema comprehension, and SQL generation, and maintain an actively updated repository of methods. However, their taxonomy does not treat reinforcement learning as a separate paradigm, nor does it cover the semantic layer or responsible AI considerations.
Shi et al.~\cite{shi2025survey} provide a comprehensive treatment of prompt engineering and fine-tuning methods with a detailed sub-categorization, published in ACM Computing Surveys. Their scope primarily covers methods through mid-2024 and does not include the recent wave of reasoning-based models (e.g., Arctic-R1, RLVR, Reasoning-SQL) or the agentic orchestration frameworks that have since redefined the state of the art.
Liu et al.~\cite{liu2024survey} adopt a lifecycle perspective spanning model design, data, evaluation, and error analysis. While their error taxonomy is a distinctive strength, they offer limited discussion of deployment-oriented topics such as semantic layers, scheduling optimization, and enterprise best practices.
Zhu et al.~\cite{zhu2024llmenhanced} survey LLM-enhanced Text-to-SQL methods with a training-strategy-based taxonomy (prompt engineering, fine-tuning, pre-trained, and agent methods), but their coverage ends before the emergence of RL-based reasoning models and does not address responsible AI or production deployment concerns.

\begin{figure}[t]
\centering
\definecolor{catcol}{HTML}{3D4555}
\definecolor{subcol}{HTML}{E8ECF2}
\definecolor{linecol}{HTML}{8A94A6}
\resizebox{\linewidth}{!}{%
\begin{forest}
  for tree={
    grow'=0,
    draw=linecol,
    rounded corners=2pt,
    align=left,
    anchor=west,
    child anchor=west,
    parent anchor=east,
    font=\footnotesize,
    inner xsep=4pt,
    inner ysep=2.5pt,
    l sep=10pt,
    s sep=3pt,
    edge={draw=linecol, thick},
  },
  forked edges,
  where level=0{fill=catcol, text=white, font=\small\bfseries, align=center}{},
  where level=1{fill=catcol, text=white, font=\footnotesize\bfseries, minimum width=3.1cm}{},
  where level=2{fill=subcol, minimum width=4.6cm}{},
  where level=3{fill=white, minimum width=7.2cm}{},
[Text-to-SQL\\Systems
  [In-Context Learning\\(Sec.~\ref{sec:icl})
    [Decomposition \& CoT prompting
      [DIN-SQL; Chain-of-Thought; Self-Consistency]]
    [Demonstration selection
      [DAIL-SQL; DCG-SQL]]
  ]
  [Reinforcement Learning\\\& Reasoning\\(Sec.~\ref{sec:rl-reasoning})
    [Policy optimization (GRPO)
      [Arctic-Text2SQL-R1; SQL-R1; Reasoning-SQL; Think2SQL]]
    [Verifiable \& preference rewards
      [Databricks RLVR-32B; ExCoT (DPO); ReViSQL]]
    [Multi-turn agentic RL
      [SQL-Trail; TRUST-SQL]]
    [Test-time scaling (hybrid)
      [Agentar-Scale-SQL]]
  ]
  [Fine-Tuning\\(Sec.~\ref{sec:finetuning})
    [Full-parameter
      [Knowledge-to-SQL; SGU-SQL; SQL-Llama (MAC-SQL);\\ CLLMs; StructLM]]
    [Parameter-efficient (PEFT)
      [FinSQL; SingSQL-LM; LoRA / QLoRA]]
    [Knowledge injection
      [Domain database knowledge injection]]
  ]
  [Agentic Methods\\(Sec.~\ref{sec:agentic})
    [Multi-agent pipelines
      [MAC-SQL; SQL-of-Thought; MAG-SQL; THOR;\\ LinkedIn KG agents; Cortex Analyst]]
    [Tool-enhanced \& execution-guided
      [CHASE-SQL; ReFoRCE; Tool-SQL; Interactive-T2S]]
    [Correction \& reliability
      [SQLFixAgent; SHARE; RTS / RTS+]]
  ]
  [RAG-Based\\(Sec.~\ref{sec:rag})
    [Table-augmented generation
      [TAG]]
    [Example retrieval \& schema pruning
      [ASTReS]]
  ]
]
\end{forest}}
\caption{Taxonomy of modern LLM-based Text-to-SQL systems covering the five learning paradigms reviewed in this survey, their main sub-categories, and representative methods. Many systems are hybrids; each is listed under its primary contribution.}
\label{fig:text2sql-taxonomy}
\end{figure}
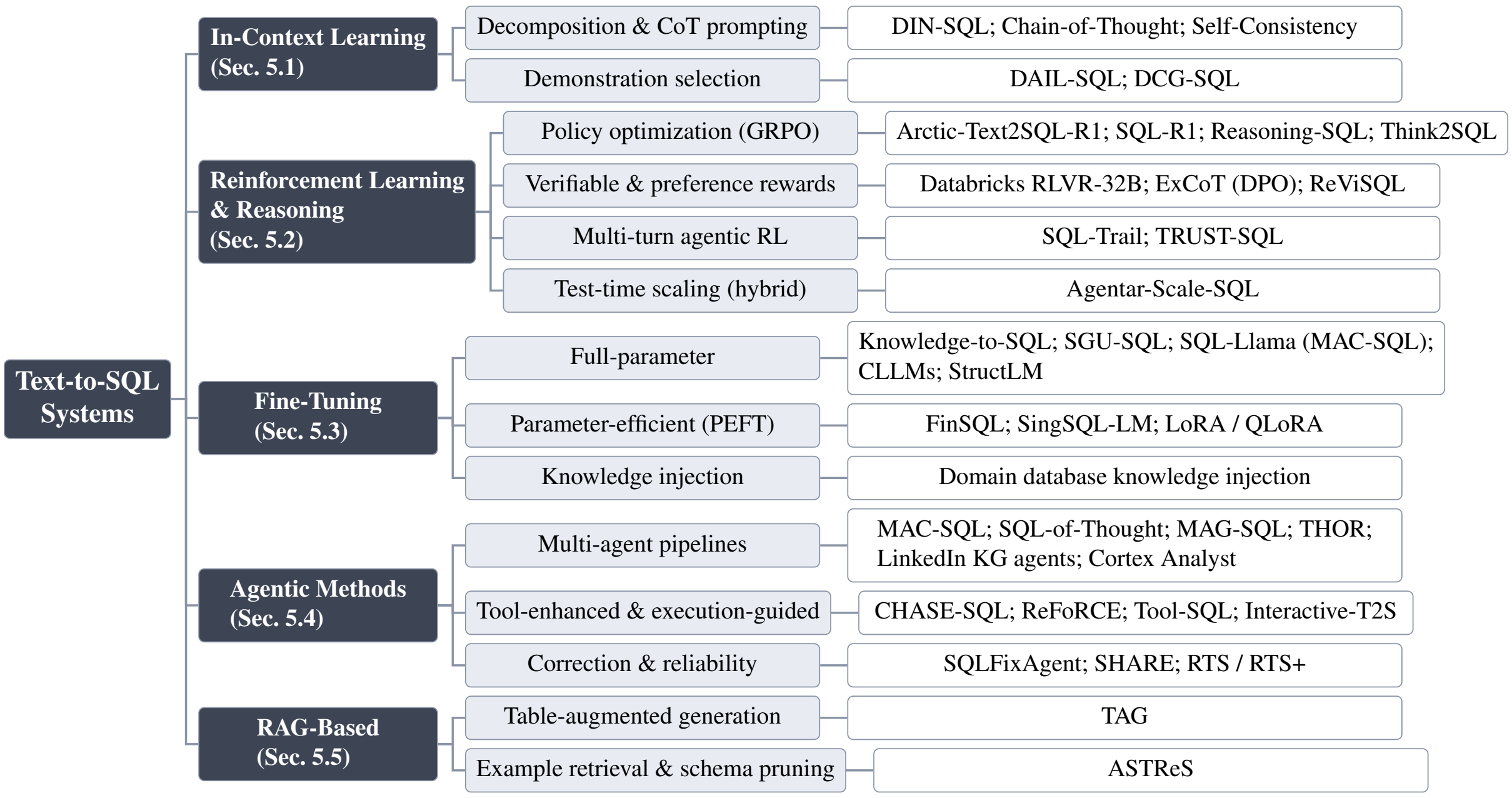

Our survey differs from and extends prior work along six dimensions:

\begin{enumerate}
    \item \textbf{Coverage of reinforcement learning and reasoning models.}
    We are the first survey to systematically cover the RL-based reasoning paradigm as a distinct methodological axis, including GRPO-based training (Arctic-R1, SQL-R1), verifiable-reward frameworks (RLVR), composite reward designs (Reasoning-SQL), DPO-based approaches (ExCoT), and the 2026 wave of data-centric and multi-turn RL methods (ReViSQL, SQL-Trail, TRUST-SQL). These methods now occupy the top positions on the BIRD leaderboard and are absent from prior surveys.

    \item \textbf{Unified five-paradigm taxonomy.}
    We propose a taxonomy that unifies approaches along five complementary paradigms (in-context learning, fine-tuning (full and parameter-efficient), agentic and tool-augmented methods, RAG-based approaches, and reinforcement learning), with explicit cross-cutting hybrid categories (Figure~\ref{fig:text2sql-taxonomy}). Prior surveys typically use two- or three-way splits (prompt engineering vs.\ fine-tuning, or ICL vs.\ fine-tuning vs.\ agents) that do not accommodate the growing diversity of hybrid methods.

    \item \textbf{Semantic layer analysis.}
    We dedicate a full section (Section~\ref{sec:semantic-layer}) to the role of the semantic layer in Text-to-SQL, covering knowledge-graph-based approaches (LinkedIn) and declarative modeling languages (Wren AI's MDL). No prior survey treats the semantic layer as a first-class architectural component, despite its importance in production deployments.

    \item \textbf{Responsible AI and ethical considerations.}
    We provide a structured discussion of privacy, transparency, bias, security, and governance (Section~\ref{sec:responsible-ai}) specific to Text-to-SQL systems that access sensitive enterprise databases. Prior surveys either omit this topic or address it only in passing.

    \item \textbf{Industry-grade deployment practices.}
    We distill actionable best practices from industry guidelines and applied research (Section~\ref{sec:best-practices}), covering caching, schema management, synthetic data generation, RAG content balancing, SQL debugging, and inference scheduling optimization (HEXGEN-TEXT2SQL). These deployment-oriented topics are not covered in prior surveys.

    \item \textbf{Recency and breadth of coverage.}
    This survey reviews over 100 papers and technical resources, with a focus on 2022--2026, including methods as recent as March 2026 (ReViSQL, TRUST-SQL) and the strongest published BIRD test result at the time of writing (Agentar-Scale-SQL, 81.67\% EX). Our coverage extends to the latest benchmarks (Spider~2.0, LiveSQLBench, BIRD-Interact) and evaluation methodologies (SQL2NL robustness, confidence estimation, SQAM/TSED), several of which postdate all prior surveys.
\end{enumerate}

The remainder of the survey is organized as follows. Section~\ref{sec:background} formalizes the task and its core challenges, Section~\ref{sec:datasets} reviews datasets and benchmarks, and Section~\ref{sec:metrics} discusses evaluation metrics. Section~\ref{sec:methodology} surveys methods across the five paradigms, and Section~\ref{sec:architecture} summarizes recurring architectural design patterns. Section~\ref{sec:semantic-layer} examines the semantic layer, Section~\ref{sec:responsible-ai} covers responsible AI, and Section~\ref{sec:best-practices} distills deployment best practices. Section~\ref{sec:conclusion} concludes with open challenges and future directions. Throughout, we highlight persistent technical challenges, including large-scale schema handling, ambiguity in natural language questions, value grounding, and cross-domain generalization, and we propose guiding principles for building safe, transparent, and dependable Text-to-SQL systems in real-world settings.

\section{Problem Definition and Background}
\label{sec:background}

This section sets up the background we rely on later in the paper. We start with a formal statement of the Text-to-SQL problem, then look at the core technical challenges that keep coming up in practice, including schema complexity, value grounding, and gaps in how these systems are evaluated.

\subsection{Text-to-SQL Task Definition}
A Text-to-SQL system takes as input a natural language question \(q = (q_1, q_2, \dots, q_n)\) and a database schema \(S = \{T_1, T_2, \dots, T_m\}\), where each table \(T_i\) comprises columns \(\{C_{i1}, C_{i2}, \dots, C_{ip}\}\). The goal is to generate an SQL query \(Q = (y_1, y_2, \dots, y_k)\) that, when executed against the relational database, returns results that precisely match the intent of \(q\):
\begin{equation}
\hat{Q} = \arg\max_{Q} \Pr(Q \mid q, S),
\end{equation}
where \(\hat{Q}\) denotes the SQL query predicted by the model.
Correctness requires both syntactic validity and semantic equivalence: the execution result of \(\hat{Q}\) must coincide with the answer implied by \(q\). Consider the following example.

\begin{center}
\small
\begin{tabular}{@{}ll@{}}
\toprule
\textbf{Column (table \texttt{Orders})} & \textbf{Description} \\
\midrule
\texttt{order\_id} & Unique ID for each order \\
\texttt{customer\_id} & ID referencing the customer \\
\texttt{order\_date} & Date of the order \\
\texttt{year} & Year extracted from \texttt{order\_date} \\
\texttt{amount} & Total monetary amount \\
\bottomrule
\end{tabular}
\end{center}

\begin{center}
\begin{tabular}{@{}l@{\quad}l@{}}
\textbf{NL:} & ``How many orders were placed in 2023?'' \\
\textbf{SQL:} & \texttt{SELECT COUNT(*) FROM Orders WHERE year = 2023;}
\end{tabular}
\end{center}

\noindent Here, the system must correctly map:
\begin{itemize}
    \item the aggregation intent ``how many'' to \texttt{COUNT(*)}, and
    \item the temporal constraint ``in 2023'' to the predicate \texttt{WHERE year = 2023}.
\end{itemize}

\subsection{Core Challenges}
Translating natural language into SQL is a complex semantic parsing problem. Key challenges include:
\begin{itemize}
  \item \textbf{Natural language understanding.} Questions may use diverse phrasings, synonyms, or colloquial expressions. Mapping ``enrolled students'' to a \texttt{Student} table and ``average score'' to \texttt{AVG(score)} requires both linguistic and domain understanding.
  \item \textbf{Schema mapping.} Schema linking requires identifying which tables and columns correspond to the entities and predicates in \(q\). Ambiguities arise when a keyword matches multiple schema elements (e.g., ``name'' could refer to \texttt{actor.name} or \texttt{director.name}) or when the vocabulary in \(q\) does not exactly match schema identifiers.
  \item \textbf{Complex query composition.} Real-world questions rarely map to a single simple query. They typically require \texttt{JOIN}s across multiple tables, nested subqueries, \texttt{GROUP BY} and \texttt{HAVING} clauses, and layered conditional logic. The space of syntactically valid SQL is enormous, and only a tiny fraction of it is semantically correct for any given question.
  \item \textbf{Multi-turn context.} In conversational interfaces, a question may depend on previous turns (e.g., ellipsis or pronoun resolution: ``How many did they order?''). Maintaining dialogue state and correctly resolving references adds a further layer of complexity.
  \item \textbf{Representative hard cases.}
    \begin{itemize}
      \item \emph{Nested aggregation:} ``Which department has more students than the average?''
      \item \emph{Implicit conditions:} ``List products that sold out last week'' (requires inferring a date range).
    \end{itemize}
  \item \textbf{Relation to semantic parsing.} Text-to-SQL sits in the broader family of semantic parsing and program synthesis, closer to translating natural language into formal logic or code than to open-ended text generation. That framing matters in practice: techniques developed for those settings, such as grammar-constrained decoding and intermediate representations like abstract syntax trees, carry over directly.
  \item \textbf{Domain generalization.} Many models degrade considerably when evaluated on unfamiliar database schemas or domains. Cross-domain generalization, i.e., understanding unfamiliar table structures, column names, and relationships, remains a key obstacle to building robust, deployable systems.
  \item \textbf{Value inference and constant prediction.} Natural language questions frequently refer to values that must be converted precisely into SQL constants (dates, integers, and strings). Mapping ``last year'' to a concrete date range or ``John'' to the correct string literal requires both contextual knowledge and dynamic value grounding against the database.
  \item \textbf{Resource constraints in deployment.} LLMs are computationally demanding and introduce latency and cost overheads in practice. Production-grade systems must optimize inference time and memory use, particularly in interactive or multi-turn settings.
\end{itemize}

\section{Datasets and Benchmarks}
\label{sec:datasets}

\begin{figure}[t]
  \centering
  \includegraphics[width=\linewidth]{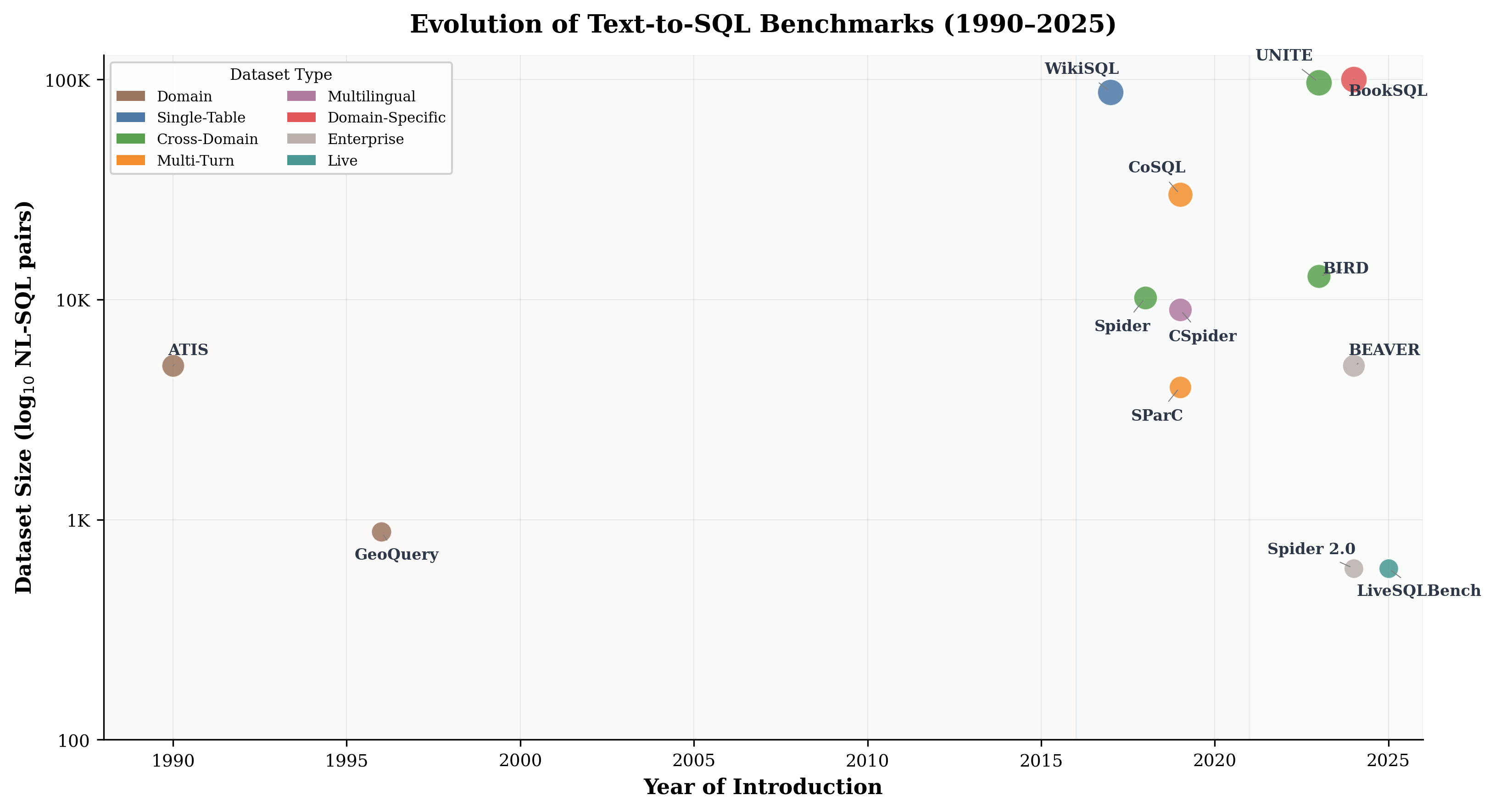}
  \caption{Evolution of Text-to-SQL benchmarks from 1990 to 2025.
  Each bubble represents a dataset positioned by its year of introduction
  (x-axis) and scale in NL--SQL pairs (y-axis, log scale), with color
  indicating dataset type. Background shading delineates three research
  eras: rule-based (pre-2016), neural (2016--2021), and LLM-driven
  (2021--present). Early benchmarks such as ATIS and GeoQuery were
  small and single-domain; the neural era introduced large-scale,
  cross-domain suites (WikiSQL, Spider, BIRD); and the LLM era has
  shifted focus toward enterprise-realistic (BEAVER, Spider~2.0) and
  contamination-free live evaluation (LiveSQLBench), prioritizing
  deployment fidelity over dataset scale alone.}
  \label{fig:dataset-timeline}
\end{figure}

The evolution of Text-to-SQL systems has been driven by the availability of increasingly large and complex benchmarks. These datasets serve both as training data and as standardized evaluation suites, enabling fair comparisons and measurable progress over time. Figure~\ref{fig:dataset-timeline} summarizes this evolution.

\subsection{Early and Domain-Specific Datasets}

Early methods relied on small, single-schema datasets that enabled the first semantic parsing experiments but lacked diversity and the complexity of real-world use cases. Common early datasets include:

\begin{itemize}
  \item \textbf{ATIS}~\cite{hemphill1990atis}: approximately 5,000 NL--SQL pairs over a flight-information schema.
  \item \textbf{GeoQuery}~\cite{zelle1996geoquery}: around 880 geographic questions over a U.S.\ geography database, originally annotated in Prolog and later converted to SQL.
  \item \textbf{Restaurants, Scholar, and Advising}~\cite{finegan2018improving}: each contains only a few hundred to a few thousand examples over a single domain schema (e.g., restaurant information or academic course advising).
\end{itemize}

\noindent\textbf{Limitation.} Small scale and a single-domain focus led to overfitting and offered little opportunity for cross-schema evaluation.

\subsection{Large-Scale Cross-Domain Benchmarks}

Modern Text-to-SQL research has centered on the following benchmarks:

\begin{itemize}

\item \textbf{WikiSQL}~\cite{zhong2017seq2sql}: 80,654 question--query pairs collected via crowdsourcing over 24,241 Wikipedia tables. Each SQL query involves a single table with simple aggregation or selection. While its scale enabled the first large-scale neural models, its restriction to single-table, non-nested queries makes it a \emph{starter} benchmark rather than a test of full SQL complexity.

\item \textbf{Spider}~\cite{yu2018spider}: 10,181 natural language questions and 5,693 unique SQL queries across 200 multi-table databases covering 138 domains. Spider includes \texttt{JOIN}s, nested subqueries, \texttt{GROUP BY}, and \texttt{HAVING}, requires generalization to unseen schemas, and is accompanied by a public leaderboard that has driven rapid innovation.

\item \textbf{Spider~2.0}~\cite{lei2024spider}: a next-generation evaluation framework of 632 real-world enterprise workflow problems that extends the canonical Spider task to \emph{end-to-end problem solving} over relational databases. Tasks are drawn from enterprise-scale databases hosted on cloud and local systems (e.g., BigQuery, Snowflake, SQLite), frequently involve schemas with more than a thousand columns, and may require multiple SQL dialects, project-level codebases, and lengthy queries. Spider~2.0 is designed to reflect realistic analytical workflows: it evaluates not only single-shot NL$\rightarrow$SQL parsers but also multi-stage pipelines and autonomous agents that inspect schemas, issue exploratory queries, validate execution, and repair errors. Two derived settings, \emph{Spider~2.0-Lite} and \emph{Spider~2.0-Snow}, isolate the pure Text-to-SQL component. Key characteristics that distinguish Spider~2.0 from earlier benchmarks include:
\begin{itemize}
  \item \textbf{Execution-centric evaluation.} Scoring is based on whether the produced results are correct when executed against the target databases, which allows single-shot parsers, prompt-based LLMs, and multi-stage agentic systems to be compared on a common footing rather than by surface-form matching.
  \item \textbf{Multi-stage workflows.} Many tasks require multi-step interaction: inspecting the schema, issuing partial queries, examining intermediate results, and performing repair or validation steps. This exposes failure modes that single-shot benchmarks cannot capture.
  \item \textbf{Tool and environment realism.} Agents operate in realistic environments with access to database metadata, documentation, and execution tools, emulating production constraints more faithfully than isolated question--SQL pairs.
  \item \textbf{Difficulty.} The benchmark is substantially harder than its predecessors: in the original study, an o1-preview-based agent solved only 17.0\% of Spider~2.0 tasks, compared with 91.2\% on Spider~1.0 and 73.0\% on BIRD.
\end{itemize}
By moving the focus from string matching to execution-guided task success and by explicitly supporting multi-stage interaction and tooling, Spider~2.0 shifts Text-to-SQL evaluation toward practical, deployment-oriented criteria, making it especially suitable for benchmarking LLM-driven agents, prompt-based systems, and hybrid pipelines intended for real analytical workloads.

\item \textbf{BIRD}~\cite{li2023llm} (BIg Bench for laRge-scale Database-grounded Text-to-SQL evaluation): a cross-domain benchmark designed to stress-test models on large, heterogeneous schemas and substantial data volumes. Here, \emph{cross-domain} refers to coverage of a wide variety of professional fields and database schemas, ranging from sports and finance to healthcare, blockchain, and education, so that evaluation is not biased toward any single domain. BIRD comprises 12,751 unique question--SQL pairs over 95 real-world databases totaling 33.4\,GB of relational data, spanning 37 professional domains, and evaluates parser robustness against schema complexity, data scale, and domain variability.

\item \textbf{UNITE}~\cite{Lan2023UNITE} (A Unified Benchmark for Text-to-SQL Evaluation): UNITE assembles and normalizes eighteen public Text-to-SQL datasets into a single unified format to evaluate generalization across diverse schemas, SQL patterns, and domains. The benchmark contains roughly 97K training and 27K test examples drawn from over 29K database instances and about 3.9K distinct SQL patterns; all constituent schemas are converted to a common SQLite representation, and splits are organized to minimize schema overlap between training and test. The authors evaluate several state-of-the-art parsers (in-context Codex, fine-tuned sequence-to-sequence models, and structure-aware variants) and report that no model exceeds roughly 50\% execution accuracy in the more realistic cross-database setting, with constrained decoding and relation-aware attention offering consistent but partial gains. UNITE therefore exposes robustness and compositionality gaps beyond Spider and serves as a diagnostic suite for future work on retrieval, schema linking, and decoding strategies.

\end{itemize}

\noindent\textbf{Comparative note.} WikiSQL tests large-scale, single-table translation with limited SQL syntax, whereas Spider and BIRD evaluate full SQL expressivity and cross-schema generalization, together covering the breadth of Text-to-SQL challenges.

\subsection{Conversational and Context-Dependent Benchmarks}

To address real-world dialogue scenarios, two conversational benchmarks extend Spider with multi-turn interactions:

\begin{itemize}
  \item \textbf{SParC}~\cite{yale2024sparc}: 4,298 coherent question sequences comprising more than 12K questions over Spider schemas, testing context retention, coreference resolution, and incremental query composition.
  \item \textbf{CoSQL}~\cite{yale2024cosql}: a large-scale, cross-domain conversational Text-to-SQL benchmark comprising over 3,000 user--system dialogues (30,000+ turns) and 10,000+ annotated SQL queries across 200 complex databases spanning 138 domains. Collected via a Wizard-of-Oz setup in which crowd workers act as users querying the database and SQL experts write the corresponding statements, CoSQL evaluates multi-turn dialogue understanding, state tracking, and dynamic SQL generation.
\end{itemize}

\noindent\textbf{Challenge.} Conversational benchmarks introduce ellipsis, anaphora, and dialogue dynamics, requiring models that encode interaction history and generate context-aware SQL.

\subsection{Emerging and Specialized Benchmarks}

Additional datasets broaden Text-to-SQL along new dimensions:

\begin{itemize}
  \item \textbf{CSpider}~\cite{Min2019PilotStudyChineseSQL} and \textbf{DuSQL}~\cite{wang2020dusql}: Chinese cross-domain Text-to-SQL datasets that test multilingual capabilities.
  \item \textbf{BookSQL}~\cite{Kumar2024BookSQL}: a large-scale, domain-specific benchmark for accounting and finance. It comprises 100,000 natural language--SQL pairs defined over a unified accounting schema that models real-world data flows such as invoices, payments, ledgers, and financial reporting. Developed in collaboration with domain experts, BookSQL emphasizes practical relevance by including query phenomena common in finance (e.g., nested aggregations, multi-table joins, and window functions). The underlying data contain approximately one million transaction records spanning 27 businesses, organized into seven relational tables.
  \item \textbf{BEAVER}~\cite{li2024beaver} (an enterprise Text-to-SQL benchmark): a dataset built from real private enterprise data warehouses, featuring hundreds of tables and thousands of columns, with natural language questions reconstructed from real query logs. In addition to anonymized schemas, the dataset provides \emph{column-mapping annotations} (phrase$\rightarrow$column) to isolate schema-linking performance. Evaluation is decomposed into table retrieval, column mapping, and end-to-end SQL generation, with execution accuracy as the primary metric; off-the-shelf LLMs perform dramatically worse on BEAVER than on public benchmarks.
  \item \textbf{Spider-DK}~\cite{gan2021spiderdk} and \textbf{Spider-Syn}~\cite{gan2021spidersyn}: Spider extensions that introduce domain-knowledge requirements and synonym substitutions, respectively, to further challenge generalization.
  \item \textbf{WikiTableQuestions}~\cite{pasupat2015wikitablequestions}: a semi-structured table QA benchmark with 22,033 hand-crafted question--answer pairs over Wikipedia tables (each with at least 8 rows and 5 columns). Unlike templated datasets, it exhibits high linguistic variability across nearly 4,000 unique column headers and many domains. Tasks include table lookup, aggregation, superlatives (argmax/argmin), arithmetic, and set operations, making it a useful testbed for Text-to-SQL systems on semi-structured data.
  \item \textbf{LiveSQLBench}~\cite{birdteam2025livesqlbench} and \textbf{BIRD-Interact}~\cite{huo2026birdinteract}: LiveSQLBench is a contamination-free, continuously updated benchmark from the BIRD team that supports the full spectrum of SQL operations (including DML and DDL) and organizes external knowledge as a hierarchical knowledge base whose entries depend on one another. BIRD-Interact builds on LiveSQLBench to evaluate \emph{interactive} Text-to-SQL: each task contains ambiguous and follow-up sub-tasks that require the system to ask clarifying questions, recover from execution errors, and adapt to evolving user requirements through a function-driven user simulator. It offers a conversational mode (c-Interact) and an agentic mode (a-Interact), with a full set of 600 tasks (up to 11,796 interactions) and a lite set of 300 tasks; in the original study, GPT-5 completed only 8.67\% of full-set tasks in c-Interact and 17.00\% in a-Interact. Including such ambiguity-resolution cases in evaluation is essential for measuring how well systems handle underspecified questions in practice.
\end{itemize}

\noindent\textbf{Summary.} From small, domain-specific collections to large, cross-domain, conversational, and interactive benchmarks, dataset evolution has been instrumental in advancing neural and LLM-based Text-to-SQL research. Despite these gains, real-world enterprise schemas, which are typically larger, more heterogeneous, and more performance-critical, remain an open frontier for future dataset development and evaluation protocols.

\section{Evaluation Metrics}
\label{sec:metrics}

\begin{figure}[t]
  \centering
  \includegraphics[width=0.78\linewidth]{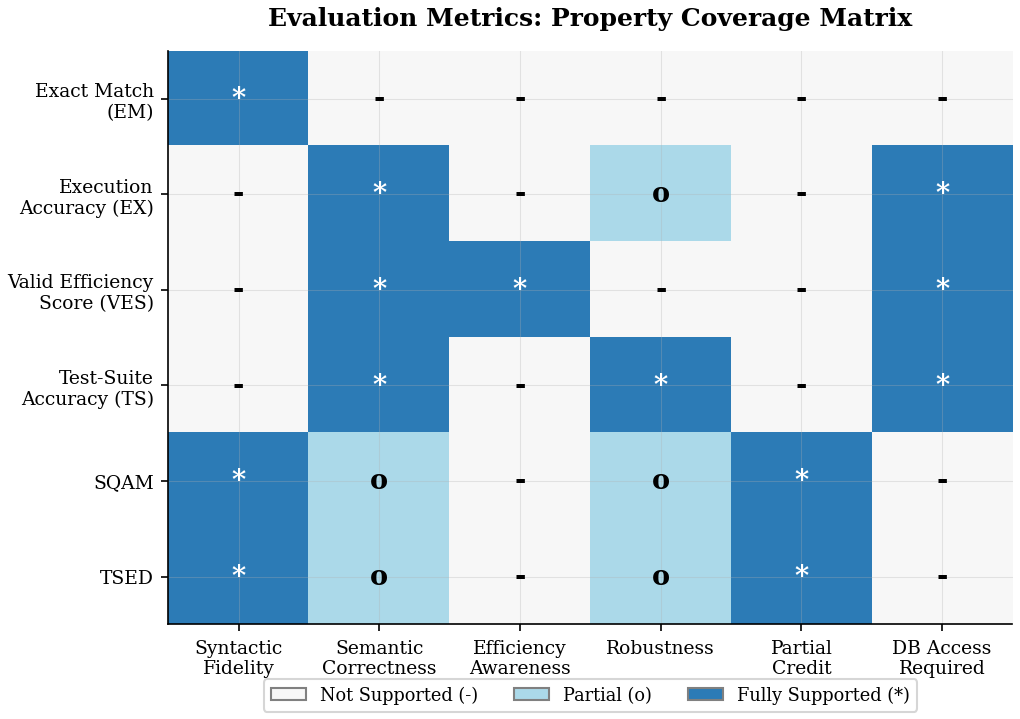}
  \caption{Property coverage matrix for Text-to-SQL evaluation metrics.
  Each cell indicates whether a metric fully supports ($*$),
  partially supports (o), or does not support (--) a
  given evaluation property. Execution Accuracy (EX) and Valid Efficiency
  Score (VES) require database access but capture semantic correctness,
  whereas structure-aware metrics such as SQAM and TSED provide partial
  credit and syntactic fidelity without requiring query execution, making
  them suitable for privacy-restricted or compliance-sensitive deployment
  settings. No single metric covers all properties, reinforcing the need
  for multi-metric evaluation protocols.}
  \label{fig:metrics-matrix}
\end{figure}

Evaluating Text-to-SQL models demands metrics that capture both the syntactic fidelity and the semantic correctness of generated queries. A robust evaluation protocol therefore verifies not only whether a predicted SQL query matches an expected syntactic form, but also whether its execution yields the same result set as the reference query. In this section we walk through the metrics that have become standard in the field, spelling out what each one actually captures and where each falls short in terms of robustness, interpretability, and how well it maps to real deployment concerns. In practice, no single metric does the job well on its own, and the right combination depends on what the system is being used for (Figure~\ref{fig:metrics-matrix}).

\subsection{Exact Match Accuracy (EM)}

\emph{Exact match accuracy} checks whether the predicted SQL query exactly matches the reference SQL query:
\begin{equation}
\text{EM} = \frac{1}{N} \sum_{i=1}^{N} \mathbb{I}(\hat{Q}_i = Q_i),
\end{equation}
where $N$ is the number of evaluation examples, $\hat{Q}_i$ is the SQL query generated by the model, $Q_i$ is the ground-truth SQL query, and $\mathbb{I}(\cdot)$ is the indicator function, which equals 1 if its argument is true and 0 otherwise.

This metric is easy to compute and interpret but is rigid: semantically equivalent queries (e.g., queries that differ only in predicate order) receive a score of 0 even though they return the same result. For example, the following two queries are semantically equivalent but fail exact match:
\begin{quote}
    \texttt{SELECT name FROM Employees WHERE dept = 'Sales' AND active = 1;}\\
    \texttt{SELECT name FROM Employees WHERE active = 1 AND dept = 'Sales';}
\end{quote}

\subsection{Execution Accuracy (EX)}

\emph{Execution accuracy} measures the correctness of SQL queries by their execution output rather than their surface form. If executing the model's query on the database yields the same result as executing the ground-truth query, the prediction is considered correct:
\begin{equation}
\text{EX} = \frac{1}{N} \sum_{i=1}^{N} \mathbb{I}\bigl(f(\hat{Q}_i, D_i) = f(Q_i, D_i)\bigr),
\end{equation}
where $f(Q, D)$ is the result of executing query $Q$ on database instance $D$, and $D_i$ denotes the database (schema and data) used for the $i$-th example.

EX captures semantic equivalence better than EM. However, it can be overly lenient: two different queries may return the same result by coincidence, especially on small databases. It also requires access to the underlying databases during evaluation.

\subsection{Evaluating Robustness via SQL2NL: Schema-Aligned Paraphrasing}

The SQL2NL framework~\cite{Safarzadeh2025SQL2NL} proposes a schema-aware approach to measure and improve the \emph{robustness} of NL-to-SQL systems under lexical and syntactic variation. Rather than paraphrasing natural language questions na\"ively (which can inadvertently change schema links), SQL2NL generates paraphrases conditioned on the \textbf{gold SQL} and the target database schema, producing diverse utterances that preserve the original intent while keeping schema alignment intact. This makes it possible to measure how sensitive a model is to wording, independently of schema-linking errors.

\paragraph{Methodology.} SQL2NL (i) ingests a gold SQL query and its schema description; (ii) prompts a generation model with constraints that encourage lexical and syntactic diversity while forbidding schema-identifier leakage; (iii) validates paraphrase fidelity through automatic embedding-based checks and targeted human review of a sample; and (iv) evaluates NL-to-SQL systems by comparing execution accuracy on the original questions with that on the SQL2NL paraphrases. The framework supports weighted (human-adjusted) scoring so that low-quality paraphrases do not unduly penalize systems. SQL2NL can also produce augmented training data for robustness-oriented fine-tuning.

\paragraph{Evaluation protocol and datasets.} The authors apply SQL2NL to Spider and BIRD and stratify their analyses by query complexity (number of joins, nested clauses, and aggregates). The primary metrics are \textbf{execution accuracy} and the \textbf{robustness gap}, defined as the drop in execution accuracy from the original to the paraphrased questions. Pass@$K$ is also reported to quantify latent capability under sampling.

\paragraph{Key results.} Experiments reveal substantial robustness deficits: large models exhibit notable execution-accuracy drops on paraphrases (around 10 percentage points for LLaMA-70B), and smaller models show even larger relative declines. Degradation increases with query complexity, with multi-join and nested queries being the most vulnerable. Pass@$K$ recovers part of the gap but is not a practical substitute for single-shot robustness in production.

\paragraph{Implications.} SQL2NL is both a diagnostic and a data-generation tool: as a diagnostic, it exposes brittle lexical generalization that standard benchmarks hide; as a data source, it provides schema-aligned paraphrases for targeted fine-tuning that mitigates this brittleness.

\paragraph{Example.}
\begin{lstlisting}
Gold SQL:
SELECT count(*) FROM orders
WHERE order_date BETWEEN '2025-06-01' AND '2025-06-30'
  AND customer_state = 'CA';

SQL2NL paraphrases (schema-aligned):
- "How many orders were placed in California during June 2025?"
- "Number of orders from customers in CA between 1 Jun 2025 and 30 Jun 2025."
\end{lstlisting}

\noindent These paraphrases preserve table and column semantics while varying the phrasing; SQL2NL measures whether a model that succeeds on the original question still returns equivalent results on these variants.

\subsection{Confidence Estimation for LLM-Based Text-to-SQL}
\label{sec:confidence}

Confidence estimation methods for Text-to-SQL~\cite{EntezariMaleki2025ConfidenceSQL} can be grouped into three pragmatic families: \textbf{black-box sampling and consistency methods}, \textbf{white-box logit aggregation}, and \textbf{execution-grounded signals}. Each family offers a different balance of accuracy, cost, and deployability.

\paragraph{Black-box sampling and consistency.}
These methods require only query access to the model and derive a reliability signal from multiple model outputs or from the model's own natural language introspection:
\begin{itemize}
  \item \textbf{Verbalized confidence.} The model is prompted to report its own confidence (single-shot or with chain-of-thought). This is easy to implement but often poorly calibrated in practice.
  \item \textbf{Consistency-based sampling.} Multiple candidate SQL queries are drawn and their agreement is measured. Practical variants include:
    \begin{itemize}
      \item \emph{Execution-based agreement:} execute the candidates and compute the fraction that return identical result sets. This is a strong signal but incurs execution cost and latency.
      \item \emph{Embedding- or clustering-based agreement:} embed the SQL queries and estimate the fraction that fall in the same cluster. This is cheaper than execution but only an indirect proxy for semantic equivalence.
      \item \emph{Schema-link agreement:} compare the extracted schema links (tables, columns, constants) across samples to estimate whether the model consistently targets the same schema elements.
    \end{itemize}
  \item \textbf{Trade-off.} Black-box consistency is straightforward to apply to closed APIs, but its quality depends on the sample size and the chosen proxy (execution is the most reliable; embedding and schema proxies are cheaper).
\end{itemize}

\paragraph{White-box logit aggregation.}
When token probabilities or logits are available, confidence can be constructed from the model's internal scores by aggregating over semantically meaningful token sets:
\begin{itemize}
  \item \textbf{Full-Token Confidence (FTC).} Aggregates probabilities over all output tokens (by averaging or multiplication). FTC is simple but treats all tokens equally and can be dominated by syntactic tokens.
  \item \textbf{Schema-Linked Confidence (SLC).} Restricts aggregation to schema-relevant tokens (table names, column names, and constants), focusing the score on the tokens that most directly affect semantic correctness.
  \item \textbf{SQL-Aware Confidence (SAC).} A syntax-aware scheme that (i) filters or down-weights non-informative tokens (formatting, aliases), (ii) normalizes semantically equivalent tokens, (iii) emphasizes critical constructs (joins, predicates), and (iv) aggregates token probabilities with a rule chosen according to query complexity. SAC further groups interchangeable constructs (e.g., commutative predicates) so that equivalent alternatives are combined before aggregation.
  \item \textbf{Trade-off.} White-box methods are generally better calibrated and more discriminative than black-box approaches, but they require access to token-level model internals and careful token selection and normalization.
\end{itemize}

\paragraph{Execution grounding and hybrid signals.}
Execution-derived signals provide direct, high-fidelity evidence of semantic correctness and can be combined with either family above:
\begin{itemize}
  \item \textbf{Single- or multi-run execution checks.} Executing one or more candidate queries and inspecting result sets or runtime errors supplies strong binary or graded confidence cues.
  \item \textbf{Hybrid fusion.} A logit-based score or sampling-consistency feature can be combined with an execution-agreement indicator to form a composite confidence score that balances immediacy and reliability.
  \item \textbf{Trade-off.} Execution grounding improves detection power at the cost of latency, compute, and the need for safe execution environments.
\end{itemize}

\paragraph{Practical guidance.}
In practice, white-box syntax-aware aggregation (SAC) is preferable when token probabilities are available and low-latency execution is infeasible; execution-based consistency is preferable when high assurance is required and execution cost is acceptable; and embedding- or schema-consistency serve as inexpensive fallbacks for closed APIs. Standard evaluation criteria for these detectors are discriminative power (AUC-ROC) and calibration quality (expected calibration error, ECE), and the final design should balance accuracy, cost, and deployment constraints.

\subsection{Valid Efficiency Score (VES)}

The \emph{valid efficiency score} jointly evaluates the correctness and execution efficiency of model-generated SQL queries~\cite{li2023llm}. Only queries whose execution result matches the ground-truth result are deemed valid; among those, more efficient executions are rewarded:
\begin{equation}
\begin{aligned}
\mathrm{VES} &= \frac{1}{N} \sum_{i=1}^{N} \mathbb{I}\bigl(f(\hat{Q}_i, D_i) = f(Q_i, D_i)\bigr) \cdot R\bigl(Q_i, \hat{Q}_i\bigr),\\
R\bigl(Q_i, \hat{Q}_i\bigr) &= \sqrt{\frac{E(Q_i)}{E(\hat{Q}_i)}}\,,
\end{aligned}
\end{equation}
where $E(Q)$ denotes the execution time of query $Q$ (or another chosen cost measure) under consistent conditions, and $R(Q_i, \hat{Q}_i)$ is the square root of the ratio of the ground-truth runtime to the generated query's runtime; the square root dampens the influence of extreme outliers.

Because only valid queries contribute and each is weighted by its relative efficiency $R$, VES captures both semantic correctness (through $\mathbb{I}$) and execution performance (through $R$). It penalizes bloated queries that are correct but inefficient. However, runtime in real deployments varies with caching and hardware, which introduces evaluation noise.

\subsection{SQAM and TSED}

To enable fair and interpretable evaluation, particularly in domains where direct database execution is restricted for privacy or compliance reasons, Song et al.~\cite{song2024enhancing} introduce two complementary structure-aware metrics: the \textbf{SQL Query Analysis Metric (SQAM)} and \textbf{Tree Similarity of Editing Distance (TSED)}. These metrics quantify both clause-level accuracy and the structural fidelity of predicted queries with respect to reference queries.

\paragraph{SQL Query Analysis Metric (SQAM).}
SQAM decomposes each SQL query into its principal clauses: \texttt{SELECT}, \texttt{FROM}, \texttt{WHERE}, \texttt{GROUP BY}, \texttt{HAVING}, and \texttt{ORDER BY}. For every clause, the overlap between the predicted and reference sub-clauses is measured with precision, recall, and a weighted F$_1$ score; semantically important clauses such as \texttt{SELECT} and \texttt{WHERE} receive higher weight in the final aggregation. In contrast to string-based metrics such as BLEU, SQAM penalizes omitted predicates, incorrect join conditions, and inappropriate aggregation choices, and therefore provides a more informative signal of partial correctness.

\paragraph{Tree Similarity of Editing Distance (TSED).}
TSED compares the abstract syntax trees (ASTs) of the generated and reference queries. Both queries are parsed into trees, and an edit distance is computed over node insertions, deletions, and substitutions; this distance is normalized by the size of the larger tree and converted into a similarity score in $[0,1]$ (e.g., $1 - \text{normalized distance}$). Because it operates on the AST, TSED captures hierarchical and structural similarity and rewards outputs that are structurally correct even when superficial token ordering or aliasing differs. Empirically, TSED correlates strongly with execution accuracy and is therefore a useful proxy when direct execution against the target database is not possible.

\paragraph{Practical significance.}
SQAM and TSED form a complementary evaluation framework: SQAM focuses on clause-level and compositional correctness, whereas TSED captures global structural alignment. Used jointly, they provide a more robust assessment of model quality than either alone. This combination is particularly valuable in industrial and financial settings, where schemas are large and complex and direct query execution may be restricted, necessitating reliable and explainable non-execution-based evaluation.

\subsection{Test-Suite Accuracy (TSA)}

\emph{Test-suite accuracy} evaluates models under more controlled and exhaustive conditions by executing each predicted query on a curated suite of databases $\mathcal{D}_i$ that share the schema of the original database but are populated to expose diverse logical edge cases. A prediction is counted as correct only if it matches the gold query's result on \emph{every} database in the suite:
\begin{equation}
\text{TSA} = \frac{1}{N} \sum_{i=1}^{N} \prod_{D \in \mathcal{D}_i} \mathbb{I}\bigl(f(\hat{Q}_i, D) = f(Q_i, D)\bigr).
\end{equation}

TSA provides a much tighter upper bound on semantic accuracy than single-database execution accuracy, because a query that is only coincidentally correct on one database instance is unlikely to remain correct across the whole suite. However, it requires generating realistic test suites and can be computationally intensive.

Additional structure-oriented metrics used in practice include:
\begin{itemize}
  \item \textbf{Component Matching (CM).} Used in the Spider benchmark, CM breaks each SQL query into its main components, including \texttt{SELECT}, \texttt{WHERE}, \texttt{GROUP BY}, \texttt{ORDER BY}, and keywords. For each component, it computes an F$_1$ score between the predicted and ground-truth sets of sub-components while ignoring order. This enables fine-grained analysis of which parts of a query are generated correctly.
  \item \textbf{Exact Set Match (EM without values).} Also used in Spider, this metric evaluates structural equivalence by treating the tables, columns, and operators within each clause as unordered sets and ignoring constant values. By focusing on structure rather than surface form, it credits semantically equivalent rewrites even when literal values differ.
\end{itemize}

\subsection{Challenges in Evaluation}

Although these metrics have helped standardize evaluation, each has limitations:
\begin{itemize}
    \item \textbf{Exact match (EM)} does not account for the syntactic flexibility of SQL and therefore penalizes queries that are logically correct but written in an alternative form.
    \item \textbf{Execution accuracy (EX)} can overestimate performance when incorrect queries coincidentally return the correct result, and it requires full access to the underlying database, which is not always feasible.
    \item \textbf{Valid efficiency score (VES)} is sensitive to execution latency and system-specific factors, limiting its reliability across environments.
    \item \textbf{AST-based similarity metrics (e.g., TSED)} incur higher computational overhead and have seen limited adoption because of their implementation complexity.
\end{itemize}

Beyond these limitations, existing metrics often fail to capture robustness to real-world language variation such as typos, paraphrasing, or underspecified questions, and they provide little insight into a model's reasoning process or its ability to recover from errors in multi-turn interactions. A further, and often overlooked, threat to validity is the quality of the benchmarks themselves: recent expert audits report annotation errors in more than half of the examined instances of BIRD Mini-Dev and Spider~2.0-Snow, enough to change both reported accuracies and leaderboard rankings~\cite{JinAnnotationErrors2026, Kim2025FLEX}. Recent work therefore increasingly favors hybrid evaluation protocols in which the semantics of the generated SQL are measured alongside syntactic faithfulness and latency, and progress in agentic methods has motivated metrics that account for tool usage and multi-step behavior.

\paragraph{Conversational quality metrics.} For multi-turn benchmarks such as SParC and CoSQL, evaluation typically includes the following measures:
\begin{itemize}
    \item \textbf{Turn-level execution accuracy:} the proportion of individual dialogue turns for which the generated SQL executes correctly, taking into account the conversational context carried over from previous turns.
    \item \textbf{Dialogue success rate:} the proportion of complete dialogues in which all turns are answered correctly, capturing the system's ability to preserve context and behave coherently throughout the interaction.
    \item \textbf{Response fluency (BLEU):} for system-generated natural language responses such as clarification questions or confirmations, BLEU or related $n$-gram overlap metrics assess fluency and similarity to reference utterances.
\end{itemize}

\paragraph{Summary.} Evaluation in Text-to-SQL has matured significantly, with EM and EX forming the core of most benchmarks. VES and TSA offer important extensions for assessing efficiency and generalization under more stringent testing. However, as models approach high accuracy on curated datasets, new metrics, as well as cleaner benchmarks, are needed to assess robustness, interpretability, and real-world deployment readiness.

\section{Methodology}
\label{sec:methodology}

This section surveys LLM-based Text-to-SQL methods across the five paradigms of our taxonomy (Figure~\ref{fig:text2sql-taxonomy}): in-context learning (Section~\ref{sec:icl}), reasoning-based and reinforcement learning approaches (Section~\ref{sec:rl-reasoning}), fine-tuning (Section~\ref{sec:finetuning}), agentic and tool-augmented methods (Section~\ref{sec:agentic}), and retrieval-augmented generation (Section~\ref{sec:rag}). Many recent systems are hybrids; we place each method according to its primary contribution and note cross-cutting elements in the text.

\subsection{In-Context Learning Approaches}
\label{sec:icl}

In-context learning (ICL) methods keep the LLM frozen and improve SQL generation through prompt design: decomposing the task into simpler steps, choosing informative demonstrations, and organizing schema information so that the model can ground the question correctly.

\paragraph{DIN-SQL.} DIN-SQL~\cite{pourreza2023din} decomposes Text-to-SQL into sub-problems that are each solved with a dedicated few-shot prompt to a frozen LLM (GPT-4 in its strongest configuration): (i) \emph{schema linking}, which identifies the relevant tables, columns, and values; (ii) \emph{classification and decomposition}, which labels each question as easy, non-nested complex, or nested complex and breaks nested questions into sub-questions; (iii) \emph{SQL generation}, which uses class-specific prompts and, for complex queries, an intermediate NatSQL~\cite{gan2021natsql} representation; and (iv) \emph{self-correction}, in which the model is prompted to review and repair its own output. Without any fine-tuning, DIN-SQL reached 85.3\% execution accuracy on the Spider test set and 55.9\% on the BIRD test set, establishing decomposed prompting as a strong baseline for later agentic designs.

\paragraph{DAIL-SQL.} Gao et al.~\cite{gao2023text} systematically benchmark prompt-engineering choices for LLM-based Text-to-SQL along three dimensions (question representation, example selection, and example organization) and distill the results into DAIL-SQL. DAIL-SQL represents the schema as SQL \texttt{CREATE TABLE} statements, selects demonstrations by combining the similarity of \emph{masked} questions (with domain-specific tokens removed) and the skeleton similarity of a preliminarily predicted SQL query, and presents examples as question--SQL pairs without their full schemas to save tokens. Combined with self-consistency, it achieved 86.6\% execution accuracy on the Spider test set with GPT-4. The same study also examines supervised fine-tuning of open-source LLMs, finding that fine-tuning substantially improves open models but can reduce their ability to benefit from in-context examples.

\paragraph{DCG-SQL.} DCG-SQL~\cite{lee2025dcgsql} (Deep Contextual Schema Link Graph) addresses demonstration retrieval for in-context Text-to-SQL by constructing a joint question--schema representation that explicitly encodes contextual links between question tokens and only the relevant schema elements. The method first prunes the schema with a RoBERTa-large cross-encoder~\cite{liu2019roberta} used as a binary relevance classifier (reported thresholds $\tau_{\text{tab}}=0.66$, $\tau_{\text{col}}=0.43$) and then converts cross-attention patterns into \emph{attention-match} edges, forming a compact contextual graph whose nodes are the pruned tables and columns and the question tokens. Each sample's graph is embedded with a relation-aware Transformer bi-encoder (220M parameters) trained contrastively with a \emph{usefulness} signal: candidates that empirically increase the likelihood of producing the anchor SQL (as measured by an LLM scorer) are treated as positives, and others as negatives. For scalability, the authors first select 100 AST-similar SQL candidates by tree edit distance and then re-rank them by usefulness to obtain hard negatives. At inference time, the user's graph is encoded, the top-$k$ demonstrations ($k=5$) are retrieved and grouped by query type (simple, join, nested), and they are injected as automated chain-of-thought templates into the generation prompt. DCG-SQL yields consistent gains across LLM backbones (e.g., on Spider, GPT-4 reaches EX\,=\,87.5 / EM\,=\,79.7 and Llama-3.1-8B reaches EX\,=\,82.1 / EM\,=\,72.9) and especially large relative improvements for smaller LLMs (+9--10\% EX and +21--23\% EM). The approach is modular and transferable, but it relies on an offline LLM-based usefulness-scoring step (mitigated by AST pre-selection) and has so far been evaluated mainly on Spider variants, leaving production-scale schema generalization to future work.

\subsection{Reasoning-Based and Reinforcement Learning Approaches}
\label{sec:rl-reasoning}

\begin{figure}[t]
  \centering
  \includegraphics[width=0.92\linewidth]{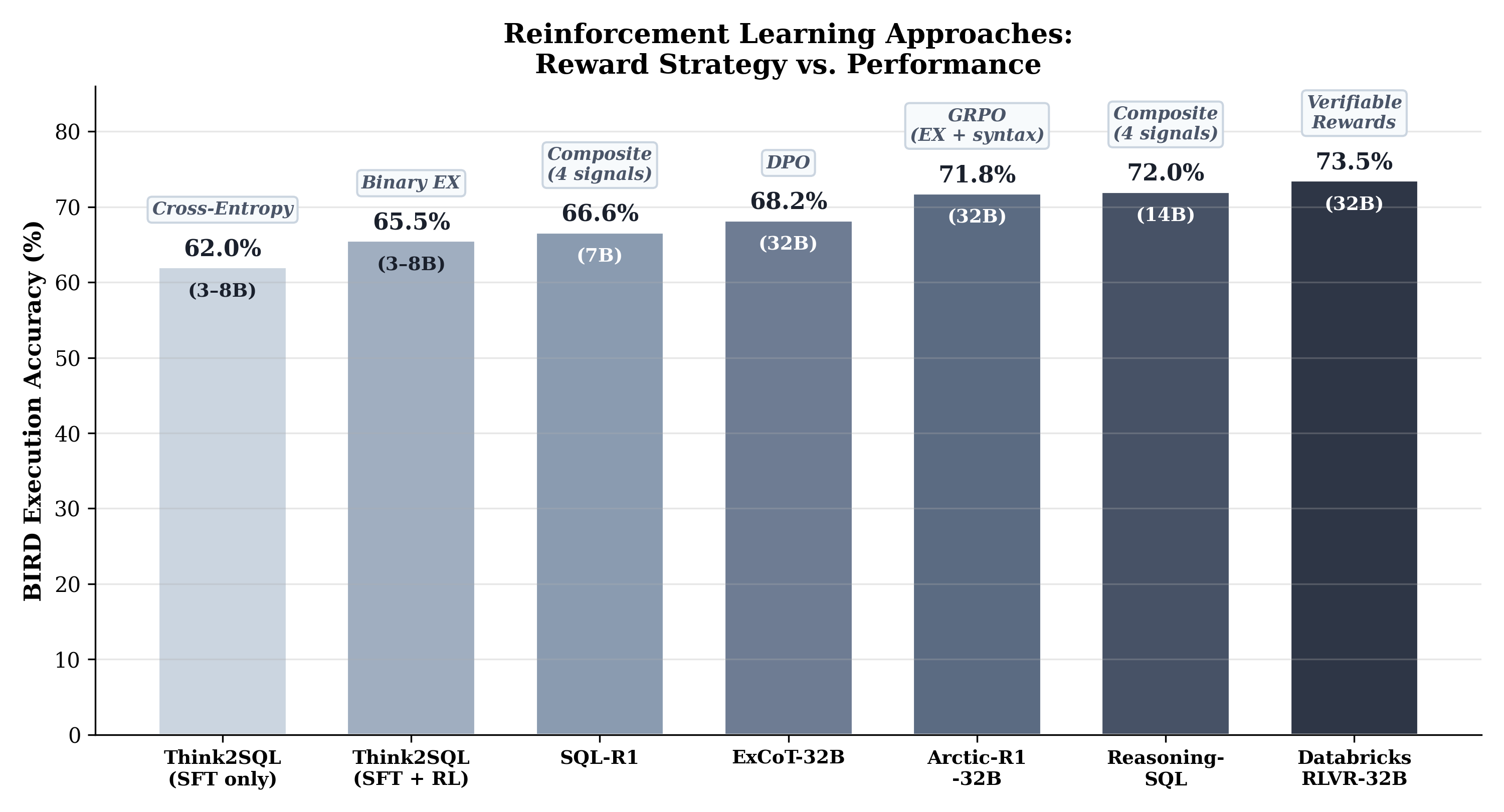}
  \caption{Comparison of reward strategies for Text-to-SQL on BIRD. Each bar shows the execution accuracy reported in the original paper, together with the model size and reward design. Values are taken as reported and mix BIRD development and test splits, so they indicate broad trends rather than a controlled comparison. Moving from supervised cross-entropy (62.0\%) to composite and execution-grounded rewards (up to 73.5\%) suggests that fine-grained or verifiable reward functions yield substantial gains over supervised training alone, even at comparable model scales.}
  \label{fig:rl-reward-comparison}
\end{figure}

\begin{figure}[t]
  \centering
  \includegraphics[width=\linewidth]{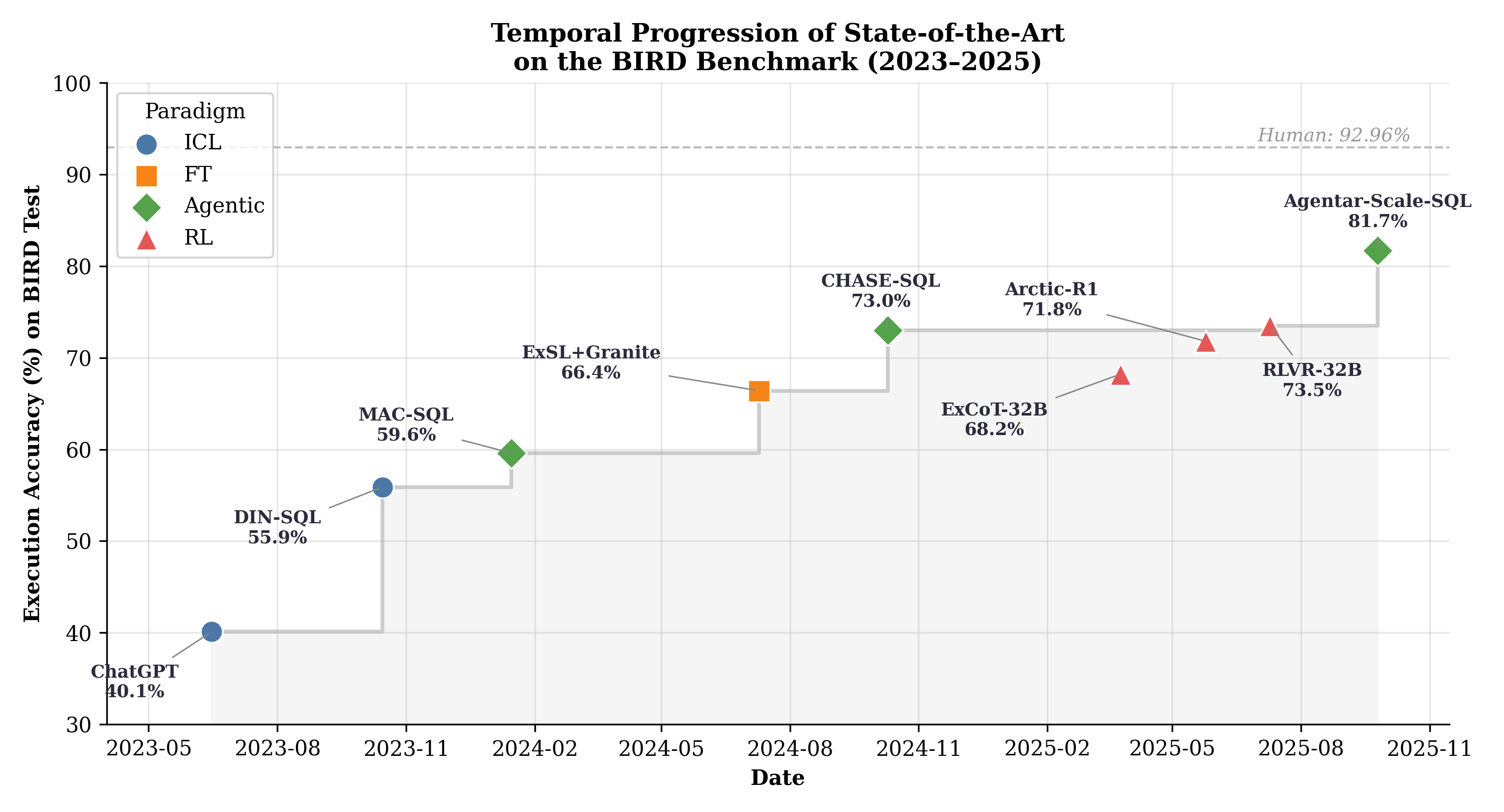}
  \caption{Temporal progression of the best reported execution accuracy (EX) on the BIRD test set (June 2023--September 2025). Each marker denotes a representative system at its time of publication, color-coded by paradigm: in-context learning (ICL), fine-tuning (FT), agentic/multi-agent, and reinforcement learning (RL). The grey frontier traces the running maximum, and the dashed line marks human expert performance (92.96\%). Over approximately two years, the best reported accuracy roughly doubled, from 40.1\% to 81.7\%, yet an 11-point gap to human performance persists on the official test set.}
  \label{fig:bird-sota-timeline}
\end{figure}

Recent work has shown that incorporating structured reasoning and execution feedback significantly improves generalization and accuracy. Reasoning-based methods model multi-step logical inference, combine intermediate thought decomposition, and use execution feedback as a learning signal, most prominently through reinforcement learning with rewards derived from executing the generated SQL. In this section, we first review the methods that established this paradigm in 2025, including Agentar-Scale-SQL, Arctic-Text2SQL-R1, and ExCoT, and then discuss three early-2026 papers that challenge its underlying assumptions. Figure~\ref{fig:rl-reward-comparison} compares representative reward designs, and Figure~\ref{fig:bird-sota-timeline} traces the resulting progress on BIRD.

\paragraph{Agentar-Scale-SQL.}
Agentar-Scale-SQL~\cite{wang2025agentarscalesqladvancingtexttosqlorchestrated} proposes an inference-time scaling framework that converts additional test-time computation into consistent accuracy gains by jointly scaling internal reasoning, sequential refinement, and parallel candidate exploration. The system combines two complementary generators: an RL-enhanced intrinsic-reasoning generator fine-tuned with execution-grounded rewards (a reward of 1 for an exact execution match and 0.1 for executable but non-matching outputs, with a KL regularizer to stabilize learning), and a retrieval-augmented ICL generator that conditions on a light schema, retrieved exemplars, and a cell-value vector database for precise value grounding. Generated queries undergo a bounded iterative refinement loop in which each candidate is executed in a sandbox, execution errors and result summaries guide targeted repairs, and repaired candidates are re-executed. Candidates are then grouped by execution equivalence and ranked by a tournament-style selector trained with reinforcement learning to prefer candidates that maximize execution accuracy; this learned selector leverages question skeletons, light-schema context, and execution evidence to outperform naive voting. The two paths are complementary: the ICL path increases coverage of rare and complex SQL skeletons, while the RL reasoning path provides higher executability and stability. The system attains state-of-the-art performance on the BIRD test set (81.67\% EX) while making the trade-off between inference cost and accuracy explicit. Gemini~2.5~Pro~\cite{google2025gemini25} serves as a backbone LLM in the pipeline.

\paragraph{Arctic-Text2SQL-R1 (Snowflake, May 2025).}
Arctic-Text2SQL-R1~\cite{yao2025arctic} reached the top of the BIRD leaderboard at the time of its release in the single-model, single-inference category: its 32B variant achieves 71.83\% execution accuracy on the BIRD test set and 70.5\% on the development set. The model adopts an execution-centric training recipe that directly optimizes for semantic correctness under database execution rather than relying on syntax-level supervision. It uses \textbf{Group Relative Policy Optimization (GRPO)}~\cite{shao2024deepseekmath}, which samples a group of SQL candidates for each question, executes them, and updates the policy according to their outcomes relative to the group mean, without a separate reward or value model. The reward is deliberately simple: 1 if the query's execution result matches the gold result, 0.1 if the query is syntactically valid and executes but returns a different result, and 0 otherwise. Combined with carefully filtered training data, strong supervised initialization (from OmniSQL checkpoints), and online training, this simple reward lets the model handle SQL aliasing and semantically equivalent variants naturally, and even the 7B variant matches earlier 70B-class systems.

\paragraph{Databricks RLVR-32B: Reinforcement Learning with Verifiable Rewards.}
Databricks Mosaic Research advanced Text-to-SQL reasoning through reinforcement learning with verifiable rewards (RLVR)~\cite{ali2025rlvrsql, databricks2025rlvr}. Unlike approaches that depend on proxy or learned reward models, RLVR exploits rewards derived directly from the correctness of SQL execution, simplifying the training pipeline and grounding supervision in the task itself. Starting from Qwen2.5-32B-Coder-Instruct~\cite{qwen2024qwen25}, the team first warm-started the model with Test-time Adaptive Optimization (TAO), an offline RL method, and then applied online RLVR training. Using no data beyond the BIRD training set and no proprietary models, the resulting model reached 73.56\% execution accuracy on the BIRD test set in the single-model, single-generation setting (surpassing the previous best single-model result of 71.83\%) and 75.68\% with self-consistency, highlighting the effectiveness of simple verifiable rewards for enterprise-grade Text-to-SQL.

\paragraph{ExCoT: Execution-Guided Chain-of-Thought (Snowflake, March 2025).}
ExCoT~\cite{zhai2025excot} combines chain-of-thought reasoning with \textbf{Direct Preference Optimization (DPO)}~\cite{rafailov2023dpo} to improve both interpretability and execution accuracy. The 32B and 70B variants of Arctic-ExCoT achieve 68.19\% and 68.53\% execution accuracy, respectively, on the BIRD test set. ExCoT uses a two-phase training pipeline. In the first (off-policy) phase, a strong LLM generates CoT--SQL pairs that are labeled positive or negative according to their execution results, and the model is fine-tuned with DPO to prefer correct outputs over incorrect ones. In the second (on-policy) phase, the updated model generates new reasoning paths, which are again validated through execution feedback and used for further rounds of preference optimization. This strategy improves compositional reasoning and executability without any handcrafted reward function, and the execution-guided iterative loop yields strong generalization to unseen database schemas.

\paragraph{Reasoning-SQL.}
Reasoning-SQL~\cite{pourreza2025reasoning} introduces a composite RL framework with \emph{SQL-tailored partial rewards} to strengthen multi-step reasoning. Recognizing that execution-only rewards are binary and sparse, it decomposes the reward into four denser components: (1) a \emph{schema-linking reward} that measures the accuracy of table and column references; (2) an \emph{AI-feedback reward} in which a separate LLM judges semantic fidelity; (3) an \emph{$n$-gram similarity reward} that encourages lexical overlap with the reference SQL; and (4) a \emph{syntax-check reward} that penalizes malformed queries. During training, multiple candidates are sampled for each question and scored by aggregating these partial rewards, producing a dense learning signal for GRPO-style group-wise optimization. A 14B-parameter model trained with Reasoning-SQL outperforms larger proprietary systems on BIRD (e.g., o3-mini by about 4 points and Gemini-1.5-Pro-002 by about 3 points) and generalizes better than purely supervised baselines, with the largest gains on complex multi-table queries.

\paragraph{SQL-R1.}
SQL-R1~\cite{ma2025sqlr1} combines supervised fine-tuning (SFT) with GRPO to train a reasoning-enhanced NL2SQL model. In the first stage, Qwen2.5-Coder-7B-Instruct is fine-tuned on \emph{SynSQL-200K}, a balanced 200,000-example subset of the 2.5M-example synthetic SynSQL-2.5M corpus that covers a range of query complexities; this ``cold start'' instills basic schema linking, SQL syntax, and the reasoning-trace format. In the second stage, GRPO is applied to \emph{SynSQL-Complex-5K}, a curated set of 5,000 challenging NL--SQL pairs requiring multi-table joins and nested subqueries. The composite reward comprises (1) a \emph{format reward} that enforces structured reasoning traces (delimited by \texttt{<think>} and \texttt{<answer>} tags), (2) an \emph{execution reward} that penalizes non-executable SQL, (3) a \emph{result reward} based on execution correctness, and (4) a \emph{length reward} that balances reasoning detail against verbosity. With only 7B parameters, SQL-R1 reports 88.6\% execution accuracy on Spider and 66.6\% on the BIRD development set, demonstrating strong data efficiency while keeping its reasoning interpretable.

\paragraph{Think2SQL.}
Think2SQL~\cite{papicchio2025think2sql} systematically studies how different training paradigms strengthen LLM reasoning for Text-to-SQL. The authors compare four approaches: (1) \emph{zero-shot learning}, in which an off-the-shelf LLM is prompted without task-specific fine-tuning (with and without generic reasoning prompts); (2) \emph{supervised fine-tuning}, in which the model is trained on annotated NL--SQL pairs with and without explicit chain-of-thought traces; (3) \emph{reinforcement learning}, using rewards such as binary execution accuracy and fine-grained metrics (precision, recall, and cardinality) that give partial credit to near-correct SQL; and (4) a \emph{two-stage SFT+RL} pipeline that first initializes the model with supervised learning and then refines it with RL. Experiments on Spider, BIRD, and two additional cross-domain benchmarks show that zero-shot prompting, even with reasoning prompts, fails on complex multi-table queries; that SFT with reasoning traces substantially improves smaller LLMs (3--8B parameters) by compensating for their limited pretraining; that pure RL yields moderate gains over SFT; and that the combined SFT+RL pipeline achieves the highest execution accuracy overall. Think2SQL thus shows that explicit reasoning supervision and fine-grained reward signals are both important for compositional generalization on challenging Text-to-SQL tasks.

\paragraph{Emerging directions (2026): rethinking the foundations of RL for Text-to-SQL.}
Three early-2026 papers each diagnose a different limitation of the RL methods above (data quality, the single-pass paradigm, and the assumption that the full schema is available), and together they suggest that closing the remaining gap to human experts may require rethinking foundational assumptions rather than scaling existing recipes.

\emph{ReViSQL (UIUC, March 2026).}
ReViSQL~\cite{Zhu2026ReViSQL} is the first method to report human-level accuracy on BIRD, reaching 93.2\% execution accuracy on an expert-verified BIRD Mini-Dev split and narrowly exceeding the 92.96\% human-expert reference reported by the original benchmark. The paper's central claim is methodological rather than architectural: it argues that the remaining gap between state-of-the-art agents and humans is driven by training-data quality rather than reasoning-pipeline design. To support this claim, the authors construct \textbf{BIRD-Verified}, a 2.5K-instance subset of the BIRD training set corrected through two to four rounds of independent review by SQL experts. They report annotation errors in 61.1\% of the audited instances (52.1\% in gold SQL queries, 26.2\% in natural language questions, and 18.2\% in external-knowledge fields), which is broadly consistent with concurrent findings on benchmark unreliability~\cite{Kim2025FLEX, JinAnnotationErrors2026}. Holding the RLVR algorithm fixed, training on BIRD-Verified instead of the original training set improves single-generation accuracy by 8.2--13.9 absolute points, showing that data quality alone accounts for a substantial share of the apparent gap. At inference time, ReViSQL samples multiple candidates from the RLVR-trained policy and uses the \emph{pre-RLVR base model} as an independent referee for execution-based reconciliation: candidates whose interpretation drifts from the user's stated intent are filtered out before a majority vote selects the final query. This mitigates the diversity collapse and over-confident ``intent fixing'' commonly observed after on-policy RL fine-tuning. The framework is released at two scales: \textbf{ReViSQL-235B-A22B}, which reaches human-level performance, and \textbf{ReViSQL-30B-A3B}, a cost-efficient Mixture-of-Experts variant that matches the prior open-source state of the art at roughly 7.5$\times$ lower per-query inference cost. ReViSQL therefore suggests that the marginal returns from rigorous data verification may now exceed those from further architectural complexity, a finding with substantial implications for how future Text-to-SQL research and engineering effort is allocated.

\emph{SQL-Trail: multi-turn RL with interleaved feedback (January 2026).}
SQL-Trail~\cite{Hua2026SQLTrail} challenges a structural assumption shared by the RL methods discussed above (Arctic-R1, RLVR-32B, ExCoT, SQL-R1, Reasoning-SQL, and ReViSQL): they all operate in a \emph{single-pass} paradigm in which the policy emits a complete SQL query in one shot and the database is consulted only afterwards to compute the reward. The authors argue that this paradigm lacks the iterative schema exploration, partial execution, and error correction that human SQL experts naturally employ, and they show empirically that single-pass RL \emph{saturates} early in training while multi-turn RL continues to improve. SQL-Trail therefore reframes Text-to-SQL as a multi-turn agentic RL problem in which each turn consists of an explicit reasoning block, an action (an executable SQL statement or a schema-inspection call), and an observation (the database's response, including errors, empty results, or sample rows). The framework rests on two contributions. First, \textbf{adaptive turn-budget allocation}: instead of a fixed interaction budget, the policy learns to scale its exploration depth to question difficulty, resolving easy questions in one or two turns and reserving longer trajectories for harder ones. This behavior emerges from a dedicated turn-number reward that penalizes unnecessary turns conditional on correctness; ablations confirm that removing it makes the model produce longer trajectories without accuracy gains, showing that the adaptive-budget property is reward-driven rather than architectural. Second, a \textbf{composite reward panel} combines (i) a final execution-correctness reward, (ii) the turn-number efficiency reward, and (iii) auxiliary shaping rewards that provide intermediate signals during the trajectory (e.g., whether a turn produced executable SQL or discovered relevant schema information). This dense reward addresses the sparsity of binary execution rewards and lets the policy learn \emph{which turn-level decisions} contributed to success. Training proceeds in two phases on a Qwen2.5-Coder backbone: a supervised cold-start phase teaches the reasoning--action--observation format, followed by GRPO-style RL with PPO-style clipped policy updates. SQL-Trail outperforms its single-pass RL counterpart trained on the same backbone and data, and on hard BIRD development questions it uses fewer turns than Claude~3.7~Sonnet while achieving higher execution accuracy. Ablations further show that removing the auxiliary shaping rewards slows convergence and lowers final performance, and that removing the SFT cold start prevents the model from learning the multi-turn format at all. Conceptually, SQL-Trail bridges the RL paradigm of this section and the agentic paradigm of Section~\ref{sec:agentic}: the iteration loops hand-engineered into systems such as MAC-SQL and CHASE-SQL can instead be \emph{learned} as part of an RL-trained policy.

\emph{TRUST-SQL: tool-integrated multi-turn RL over unknown schemas (March 2026).}
TRUST-SQL~\cite{Jian2026TRUSTSQL} extends the multi-turn RL paradigm of SQL-Trail to a more realistic deployment setting that exposes a hidden assumption shared by essentially every method in this survey: the \emph{Full Schema Assumption}, whereby the complete database schema is injected into the model's context alongside the question. Real enterprise databases contain hundreds to thousands of tables with voluminous and noisy metadata, making upfront schema injection infeasible. TRUST-SQL therefore studies the \emph{Unknown Schema} scenario, in which the agent starts with no schema knowledge and must actively discover, inspect, and verify the relevant subset before generating any SQL. The task is formulated as a partially observable Markov decision process whose hidden state is the full schema and whose observations are limited to what the agent's metadata-inspection tools return. On top of this formulation, TRUST-SQL imposes a \textbf{four-phase interaction protocol}: \emph{Explore} (schema-discovery tool calls such as table listing, column inspection, and row sampling), \emph{Propose} (explicit commitment to a verified subset of tables, columns, and join paths), \emph{Generate} (SQL conditioned on the proposed schema), and \emph{Confirm} (execution and validation of the result). The Propose phase acts as a \emph{mandatory cognitive checkpoint}: in a pilot study, schema hallucinations (references to tables or columns that do not exist) account for 26.4\% of failures in a vanilla execution-correction baseline, 14.2\% when execution feedback alone is added, and only 2.8\% once the Propose checkpoint is introduced (a 9.4-fold reduction).

The protocol also provides the structural boundary required by the paper's second contribution, \textbf{Dual-Track GRPO}. The authors identify a credit-assignment failure in standard GRPO on heterogeneous multi-turn trajectories: because GRPO assigns the same group-relative advantage to every token in a trajectory, a failure during SQL generation wrongly penalizes tokens emitted during earlier schema exploration, and vice versa. Dual-Track GRPO decomposes each trajectory into a \emph{schema track} (tokens generated during Explore and Propose) and an \emph{execution track} (tokens generated during Generate and Confirm) and applies \emph{token-level masked advantages}, so that each token is updated only by the reward associated with its phase. Schema-track tokens are optimized against an exploration-grounded reward that measures whether the proposed schema subset matches the one required by the gold SQL, while execution-track tokens are optimized against the usual execution-correctness reward. Holding the protocol, backbone, and training data constant, Dual-Track GRPO yields a 9.9\% relative improvement over standard GRPO. Training uses Qwen3-4B and Qwen3-8B backbones, with a supervised cold-start phase followed by Dual-Track GRPO with PPO-style clipped updates. Across five Text-to-SQL benchmarks, TRUST-SQL delivers average absolute execution-accuracy improvements of \textbf{30.6} points for the 4B variant and \textbf{16.6} points for the 8B variant over their respective base models; the larger gain for the smaller model suggests that structured agentic protocols benefit compact models disproportionately. Most importantly for deployment, TRUST-SQL operates entirely \emph{without a preloaded schema} yet consistently matches or surpasses strong baselines that rely on full schema injection.

\subsection{Fine-Tuning Approaches}
\label{sec:finetuning}

Fine-tuning remains a cornerstone for adapting pretrained models to Text-to-SQL. Depending on whether all parameters or only a subset are updated, fine-tuning methods fall into two categories: full-parameter fine-tuning and parameter-efficient fine-tuning. We discuss representative models in each category and then a data-centric knowledge-injection approach that precedes task fine-tuning.

\paragraph{Full-parameter fine-tuning.}
In full-parameter fine-tuning, every weight of the pretrained LLM is updated on task-specific data. This approach typically yields the highest accuracy on complex benchmarks but incurs significant computation and memory overhead.

\begin{itemize}
  \item \textbf{Knowledge-to-SQL}~\cite{hong2024knowledge} fine-tunes a \emph{Data Expert LLM} (DELL) that generates the domain knowledge needed to answer a question (for example, the definition of a metric or the meaning of a coded value) and passes it to a downstream Text-to-SQL model. DELL is first trained with supervised fine-tuning and then refined with Direct Preference Optimization using database execution feedback: knowledge that leads the downstream model to correct SQL is preferred over knowledge that does not. Because it supplies knowledge rather than SQL, DELL can be paired with different Text-to-SQL models and improves their execution accuracy on BIRD and Spider.

  \item \textbf{SGU-SQL}~\cite{zhang2024structure} constructs dual graph embeddings: one for the database schema (tables as nodes, foreign keys as edges) and one for the query's grammar tree. A node-level alignment process propagates contextual information across the merged schema--query graph to map question tokens precisely to schema elements, and an auxiliary structure-consistency objective encourages agreement between the schema and query graphs. SGU-SQL improves multi-table join and nested-query generation on Spider and BIRD.

  \item \textbf{SQL-Llama (MAC-SQL)}~\cite{wang2023macsql}. Alongside its GPT-4-based multi-agent framework (Section~\ref{sec:agentic}), MAC-SQL releases SQL-Llama, an open-source model obtained by instruction-tuning Code Llama-7B~\cite{roziere2024codellama} on agent-specific instruction data covering the Selector, Decomposer, and Refiner roles. A single fine-tuned model can then play all three agents, recovering much of the GPT-4 variant's accuracy with a far smaller, locally deployable model.

  \item \textbf{CLLMs}~\cite{kou2024cllms} (Consistency Large Language Models) target inference efficiency rather than accuracy. They fine-tune an LLM so that it maps any intermediate state of a Jacobi (parallel) decoding trajectory directly to the final fixed point, allowing several tokens to be generated per step. On Text-to-SQL, a DeepSeek-Coder-7B model fine-tuned on Spider with this consistency objective decodes several times faster while preserving generation quality, illustrating how fine-tuning can reduce latency in interactive Text-to-SQL deployments.

  \item \textbf{StructLM}~\cite{zhuang2024structlm} is a generalist model for structured knowledge grounding. It fully fine-tunes CodeLlama-Instruct models (7B to 34B parameters) on SKGInstruct, an instruction-tuning dataset of roughly 1.1M examples spanning 18 tasks over tables, databases, and knowledge graphs, including Text-to-SQL datasets such as Spider and CoSQL. StructLM outperforms task-specific models on the majority of the evaluated datasets, and the authors observe that scaling from 7B to 34B brings only marginal gains, suggesting that structure-aware training data matters more than model size for this family of tasks.
\end{itemize}

\paragraph{Parameter-efficient fine-tuning.}
Parameter-efficient fine-tuning (PEFT) updates only a small subset of weights (e.g., adapter modules, low-rank matrices~\cite{hu2021lora, dettmers2023qlora}, or output heads) while freezing the majority of pretrained parameters. This reduces GPU memory use and training time while preserving most of the original model's linguistic generalization, which makes it attractive for domain-specific and privacy-sensitive deployments.

\begin{itemize}
  \item \textbf{FinSQL}~\cite{guo2024finsql} is a model-agnostic, privacy-aware framework tailored to financial analysis. The authors release BULL, a bilingual (English--Chinese) benchmark of 4,966 NL--SQL pairs over fund, stock, and macroeconomic databases characterized by complex schemas and domain-specific abbreviations. FinSQL combines prompt construction with hybrid data augmentation, LoRA-based multi-task PEFT on open-source LLMs, and output calibration, avoiding reliance on proprietary APIs. It achieves up to 36.6\% improvement over baselines in few-shot cross-database transfer, offering a scalable and secure option for sensitive financial environments.
  \item \textbf{SingSQL-LM}~\cite{caferoglu2025singsql} fine-tunes compact Qwen2.5-Coder models (1.5B and 3B parameters) with LoRA on in-domain synthetic data generated for a single target database (Section~\ref{sec:best-practices}), illustrating how PEFT and synthetic data together enable inexpensive schema-specialized models.
\end{itemize}

\paragraph{Domain knowledge injection.}
Ma et al.~\cite{Ma2024DomainKE} propose a data-centric fine-tuning pipeline that injects compact domain database knowledge (schema tokens and sampled cell values) into an LLM to reduce column and table hallucinations and value--column mismatches. By converting the schema and sampled values into targeted generation tasks, the model internalizes value-to-column and column-to-table associations and learns schema co-occurrence patterns. This knowledge-injection stage precedes standard Text-to-SQL fine-tuning and yields consistent improvements in execution accuracy and exact-match fidelity across several open LLMs, while also increasing robustness to lexical variation.
\begin{itemize}
  \item \textbf{Problem.} LLMs often hallucinate column or table names or misplace values because they lack grounding in the actual database contents.
  \item \textbf{Core idea.} Inject the schema and representative cell values during training so that models learn value--column and column--table associations.
  \item \textbf{Training objectives.} Three compact tasks: (i) column semantic enrichment (value to column), (ii) table inference (row or column set to table), and (iii) co-occurrence and join feasibility (column grouping and join prediction).
  \item \textbf{Sampling.} Rows are clustered, high-quality rows are sampled from each cluster, and the proportion of numeric values is controlled to build informative, compact sub-tables.
  \item \textbf{Workflow.} Two-phase training: knowledge-injection training followed by downstream Text-to-SQL fine-tuning.
  \item \textbf{Results.} Consistent gains in execution accuracy and exact match on Spider variants, and improved robustness on a synonym-substituted development set.
  \item \textbf{Limitations.} The approach requires moderate engineering effort for data extraction and sampling and may need adaptation for unusual enterprise schemas.
\end{itemize}

\subsection{Agentic Methods}
\label{sec:agentic}

Agent-based approaches have emerged as a powerful paradigm for improving semantic alignment, modular interpretability, and execution correctness in complex query generation. By coordinating multiple LLM-driven agents and external tools, these systems decompose, validate, and refine SQL queries with higher accuracy and robustness than monolithic architectures. This section surveys a range of recent agentic frameworks that exemplify the shift toward collaborative, multi-step processing.

\paragraph{ReFoRCE.}
ReFoRCE~\cite{deng2025reforce} addresses the practical failure modes of large-scale, cross-domain Text-to-SQL by combining schema compression, execution-guided self-refinement, and selective column exploration in a single agentic pipeline. The system first reduces prompt bloat by grouping similarly named tables by pattern and retaining representative table DDLs and sample rows, and then runs a lightweight LLM-driven schema linker to select task-relevant tables and columns. Using the compressed context, the pipeline samples multiple SQL candidates, executes them in a sandbox, and invokes an LLM-based self-refinement loop that repairs syntax, dialect, and semantic errors. Candidates are aggregated by majority vote, and low-confidence cases trigger targeted, multi-threaded column probes whose sample values are fed back to the model for disambiguation. Ablations attribute most of the gains to the compression stage, and the combined generate--execute--refine--explore workflow achieved state-of-the-art results on Spider~2.0 (approximately 35.8\% execution accuracy on the Snow split and 36.6\% on the Lite split, with substantial EX@$k$ gains), at the cost of a modest number of additional LLM and database calls per example. ReFoRCE thus illustrates a practical trade-off between prompt tractability, execution grounding, and inference cost for large, heterogeneous schemas.

\paragraph{Cortex Analyst (Snowflake).}
Cortex Analyst~\cite{snowflake_cortex_analyst2024} is Snowflake's enterprise-grade agentic framework for Text-to-SQL, designed to let non-technical users interact with structured data. At its core is a semantic model encoded in YAML that captures domain-specific table and column metadata, relationships, and synonyms. An orchestrator coordinates several specialized components: a relationships agent infers join paths, a semantic-model editor updates metadata using execution traces, a custom-instruction agent injects domain-specific behavior, and an evaluator agent performs multi-step validation. SQL generation is guided by Claude~3.5~Sonnet and refined through an execution-feedback loop. Snowflake reports roughly 90\% execution accuracy on real-world business-intelligence datasets, an improvement of about 20--30\% over LLM-only approaches.

\paragraph{SHARE.}
SHARE~\cite{SHARE2025} (an SLM-based Hierarchical Action corREction assistant) proposes a lightweight, modular self-correction pipeline that replaces costly recursive LLM repair loops with three specialized small language models (SLMs, each under 8B parameters): a \emph{Base Action Model} (BAM) that converts an initial SQL candidate into an explicit \emph{action trajectory} (a stepwise reasoning trace), a \emph{Schema Augmentation Model} (SAM) that refines schema bindings and table/column alignments within the trajectory, and a \emph{Logic Optimization Model} (LOM) that corrects semantic and logical defects before the final SQL is reconstructed. The components are trained with a \emph{hierarchical self-evolution} scheme in which each SLM is progressively specialized with compact, task-specific supervision, so as to localize errors precisely and improve repair accuracy with minimal data. By operating on intermediate action trajectories rather than raw SQL text, SHARE achieves fine-grained error localization and substantially reduces end-to-end compute compared with monolithic LLM recursion. It yields large single-iteration execution gains (+14.8\% on BIRD and +11.4\% on Spider) while remaining compatible as a post-hoc corrector for a variety of generator LLMs. The framework is currently evaluated in a single-turn correction setting and is specialized to Text-to-SQL, leaving multi-turn interactive correction and broader code-generation settings to future work.

\paragraph{RTS / RTS+: Reliable Text-to-SQL with adaptive abstention.}
Unlike most agentic frameworks, which assume that the system must always produce a SQL query, RTS~\cite{Chen2025RTS, Chen2025RTSplus} targets deployment scenarios in which a confidently wrong query is more harmful than an abstention, a setting common in enterprise applications such as finance and healthcare. The framework concentrates its reliability machinery on the schema-linking phase, which the authors identify empirically as the dominant source of downstream errors. Its central technical contribution is \textbf{Branching Point Prediction (BPP)}, a mechanism that inspects the LLM's hidden-layer representations during schema-linking token generation to identify positions at which the model is uncertain enough that an alternative token could change the final SQL. Crucially, BPP applies \emph{conformal prediction}~\cite{vovk2005algorithmic, papadopoulos2002conformal, barber2023conformal} to these uncertainty signals, providing formal statistical guarantees that the flagged branching points cover the true points of ambiguity at a user-chosen risk level $\alpha$. This places RTS in a different epistemic category from the heuristic confidence estimators discussed in Section~\ref{sec:confidence} (e.g., FTC, SLC, and SAC), which provide calibrated scores but no coverage guarantees. When BPP fires, the system either abstains or asks the user a targeted clarifying question, naturally instantiating a human-in-the-loop workflow. On BIRD, RTS achieves near-perfect schema-linking accuracy with bounded human intervention, and when combined with a small downstream SQL generator it nearly matches the execution accuracy of state-of-the-art systems whose generators are orders of magnitude larger, suggesting that selective abstention plus a small model is a competitive alternative to scaling. The companion system \textbf{RTS+}~\cite{Chen2025RTSplus} packages the framework as an interactive demonstration with a graphical interface for schema-linking decisions, query verification, refinement, and abstention analytics, and reports that users who answer a small number of clarification prompts can achieve perfect scores on selected query sets. RTS thus exemplifies an under-explored design point: prioritizing calibrated uncertainty and mixed-initiative interaction over fully autonomous generation, with direct implications for trustworthy enterprise deployment.

\paragraph{MAC-SQL.}
MAC-SQL~\cite{wang2023macsql} proposes a multi-agent collaborative pipeline comprising three agents: a \emph{Selector}, a \emph{Decomposer}, and a \emph{Refiner}. The Selector filters the schema to retain only relevant tables and columns, reducing token overhead; the Decomposer applies chain-of-thought reasoning to break complex questions into sub-problems; and the Refiner executes candidate queries, collects errors or empty results, and iteratively fixes the outputs using this feedback. The framework is instantiated with GPT-4 for maximum accuracy and with SQL-Llama (Section~\ref{sec:finetuning}) for open-source deployment. MAC-SQL set a new state of the art on BIRD at the time of publication (59.59\% execution accuracy on the test set) and is competitive on Spider; its ablations show that each agent contributes to performance, confirming the value of multi-agent collaboration and error recovery.

\paragraph{SQL-of-Thought.}
SQL-of-Thought~\cite{Chaturvedi2025SQLofThought} presents a multi-agent pipeline that couples staged chain-of-thought planning with a taxonomy-guided correction loop to reduce logical and schema-related errors. Generation is decomposed across specialized agents (schema linking, sub-problem decomposition, query planning, and SQL synthesis). Upon execution failure, a correction stage driven by a fine-grained error taxonomy diagnoses the fault (e.g., join mistakes, schema-linking errors, or aggregation misuse) and issues a focused repair plan; this structured feedback replaces undirected regeneration and yields more effective repairs. Architecturally, the system enforces a plan-first workflow in which procedural query plans precede SQL synthesis, uses constrained decoding and post-execution validation, and repeats correction attempts until success or a fixed budget is reached. The approach attains state-of-the-art execution accuracy on the Spider development set (91.6\%) and strong robustness on Spider-Realistic (90.2\%) and Spider-Syn (82.0\%), with a very high rate of valid SQL; ablations show that removing either the query-planning stage or the taxonomy-guided correction loop causes substantial accuracy drops. The authors also characterize practical trade-offs: multi-agent planning and iterative correction substantially increase token and latency costs, and they suggest hybrid deployments (stronger models for planning, cheaper models for generation) to balance cost and performance. SQL-of-Thought is therefore particularly attractive when correctness and auditability are priorities, albeit at higher compute and engineering cost.

\paragraph{CHASE-SQL: multi-path reasoning and candidate selection.}
Traditional Text-to-SQL systems often rely on single-shot generation, which limits their robustness on complex queries. CHASE-SQL~\cite{pourreza2024chase} instead combines diverse reasoning strategies with preference-optimized candidate selection. It employs three complementary candidate generators: divide-and-conquer decomposition of the question, query-plan-guided chain-of-thought reasoning that mirrors the steps of a database engine, and online synthetic example generation that produces instance-aware few-shot demonstrations tailored to the test question. A fine-tuned selection agent then compares candidates pairwise to identify the best query, and the pipeline uses locality-sensitive hashing (LSH)~\cite{indyk1998lsh} for retrieving relevant database values together with a query fixer to repair syntactic errors. CHASE-SQL achieved a new state-of-the-art execution accuracy of 73.01\% on the BIRD test set at the time of publication, substantially outperforming conventional self-consistency and ranking-based selection and highlighting the effectiveness of multi-path reasoning combined with learned selection.

\paragraph{Text-to-SQL for enterprise data analytics (LinkedIn).}
Chen et al.~\cite{Chen2025EnterpriseSQL} describe LinkedIn's enterprise-scale Text-to-SQL system, which combines a dynamic knowledge graph with a multi-agent orchestration framework to support reliable natural language querying over large and continuously evolving databases. The knowledge graph integrates heterogeneous enterprise artifacts (table and column schemas, internal documentation, historical query logs, code repositories, and crowdsourced domain notes) into a unified semantic representation (Section~\ref{sec:semantic-layer}). Each table node carries metadata such as certification status, usage frequency, popularity signals, and representative values, and user access patterns are clustered with independent component analysis (ICA) to enable personalized, context-aware schema retrieval. In the multi-agent pipeline, a supervisor agent first identifies user intent and triggers a \emph{knowledge-graph retriever} that selects relevant context. A \emph{query-writer} agent then generates candidate SQL from the retrieved schema information and examples, after which a \emph{researcher} agent verifies schema usage and cross-checks the generated queries against documentation to reduce hallucinations. Validated queries are executed under read-only constraints, and results are presented to users together with provenance information and optional clarifications. Expert evaluation on internal benchmarks reports that approximately 53\% of responses are correct or close to correct, and ablations highlight the importance of knowledge-graph-driven personalization and targeted validation. The system illustrates that robust industrial Text-to-SQL benefits from combining structured enterprise knowledge with coordinated agent-based reasoning rather than relying on standalone LLM inference.

\paragraph{SQLFixAgent.}
SQLFixAgent~\cite{cen2024sqlfixagent} improves semantic accuracy through a three-agent architecture consisting of \emph{SQLReviewer}, \emph{QueryCrafter}, and \emph{SQLRefiner}. SQLReviewer identifies syntactic and semantic errors using a simulated rubber-duck debugging strategy; QueryCrafter produces a diverse set of candidate repairs with a fine-tuned SQL generation model; and SQLRefiner selects the most appropriate query by leveraging a failure memory and similarity-based retrieval over past corrections. On BIRD, SQLFixAgent improves on prior methods by approximately 3\%, achieves an executable-SQL rate of 89\%, and reduces semantic error rates by 41\%. The framework is model-agnostic and supports plug-and-play integration with LLMs such as GPT-4 and CodeLlama.

\paragraph{THOR.}
THOR (Transformer Heuristics for On-Demand Retrieval)~\cite{Shi2025THOR} is a production-focused Text-to-SQL system for safe, read-only analytical querying over enterprise databases. Its architecture decouples orchestration from execution: a supervisor routes user requests, a schema-retrieval agent supplies a targeted subset of the schema, and a constrained SQL-generation agent produces single-statement \texttt{SELECT} queries under strict read-only guardrails. A self-correction and rating module analyzes execution failures and applies targeted fixes through an iterative diagnose--plan--regenerate loop, after which a result-interpretation agent translates validated outputs into human-readable insights. Central design choices include on-demand schema injection to limit hallucinations, constrained generation to enforce safety policies, and a directed retry mechanism to improve execution success. Pilot studies across finance, sales, and operations workloads report notable gains in task success and reliability, albeit with higher token usage and latency due to iterative refinement, indicating that THOR suits enterprise scenarios in which correctness, safety, and auditability take precedence over minimal response time.

\paragraph{Tool-SQL.}
Tool-SQL~\cite{Wang2024ToolAssistedAgent} introduces tool-assisted, clause-level validation of LLM-generated SQL. Each query is produced as a sequence of function calls (e.g., \texttt{add\_where(col, op, val)}) and then checked by two tools: a \emph{database retriever} that verifies value--cell alignment and an \emph{error detector} that identifies silent mismatches and constraint violations not caught by execution. Iterative refinement is guided by tool feedback until the query is accepted or a retry limit is reached. The authors also introduce Spider-Mismatch, a benchmark containing realistic mismatch errors. Tool-SQL achieves top-tier performance on Spider and Spider-Realistic in few-shot settings and excels at ensuring logical correctness beyond mere execution success.

\paragraph{MAG-SQL.}
MAG-SQL~\cite{Xie2024MAGSQL} is a modular multi-agent generative framework for Text-to-SQL over large schemas. It introduces \emph{soft schema linking}, in which all columns are retained in the prompt but enriched selectively, according to entity matches with the question, with table descriptions, examples, and definitions. Complex questions are decomposed by a \emph{targets--conditions decomposer}, which first identifies the core query intent and then incrementally introduces filtering and constraint conditions. Sub-SQL statements are generated sequentially with chain-of-thought-style prompting, keeping intermediate reasoning explicit, and a \emph{sub-SQL refiner} executes each intermediate query and applies targeted corrections in response to execution errors or unexpected results. This iterative generate--execute--refine process improves accuracy: the GPT-4-based variant reaches 61.08\% execution accuracy on BIRD and shows notable improvements in robustness, transparency, and error recovery on complex multi-hop questions.

\paragraph{Interactive-T2S.}
Interactive-T2S~\cite{InteractiveT2S2024} recasts SQL generation as a multi-turn, tool-mediated process rather than a one-shot generation task. The framework defines four general-purpose tools: (i) a \emph{schema-linking tool} that prunes large schemas to the tables and columns relevant to the question; (ii) a \emph{cell-value localization tool} that grounds user-specified values such as dates and locations in concrete database entries to reduce hallucinations; (iii) a \emph{table-joining tool} that discovers valid foreign-key relations and enforces semantically consistent joins; and (iv) a \emph{query-refinement tool} that uses execution feedback to correct and optimize candidate SQL. The method follows a staged pipeline: the schema is narrowed by linking, candidate values are localized, valid join paths are determined, and a candidate query is executed and refined in a bounded loop. This modular decomposition improves interpretability and mitigates the inefficiencies of wide-schema settings in which single-pass approaches often fail. On BIRD-Dev, the system attains state-of-the-art performance in the setting without oracle knowledge while using only two few-shot examples, indicating that stepwise tool use is a sample-efficient, interpretable, and scalable strategy for high-accuracy Text-to-SQL.

\subsection{RAG-Based Approaches}
\label{sec:rag}

\paragraph{TAG.}
Table-Augmented Generation (TAG)~\cite{biswal2024tag} is a unified framework that extends beyond traditional Text-to-SQL and retrieval-augmented generation by integrating table-centric reasoning with the world knowledge and reasoning capabilities of LMs. The authors observe that many existing systems translate language to SQL accurately at a syntactic level but struggle with questions that require semantic inference, world knowledge, or nuanced interpretation. The TAG pipeline consists of three interlinked steps: (1) \emph{query synthesis}, in which an LM translates the user request $R$ into an executable query $Q$ (typically SQL); (2) \emph{query execution}, in which $Q$ is run against the database to produce intermediate table results $T$; and (3) \emph{answer generation}, in which the LM conditions on both $R$ and $T$ to produce the final natural language answer $A$. This stepwise composition decouples logical-form prediction from answer rendering, enabling recursive reasoning and richer interpretability. To assess TAG, the authors introduce \textbf{TAG-Bench}, derived from a modified BIRD corpus and comprising 80 challenging queries that span match-based retrieval, comparisons, aggregations, and ranking; roughly half require external world knowledge, and the rest demand reasoning beyond schema-level inference. Baselines such as conventional Text-to-SQL, retrieval-augmented generation, and LM re-ranking all perform poorly on TAG-Bench, with exact-match accuracies of at most 20\%, whereas hand-written TAG pipelines implemented on the LOTUS runtime achieve absolute gains of roughly 20--65 percentage points. These results indicate that table-augmented generation substantially improves robustness on complex analytical queries, and the open-source release supports reproducibility.

\paragraph{ASTReS.}
ASTReS~\cite{ASTReS2024} (AST-based REranking and Schema pruning) is a retrieval-augmented framework designed to overcome the shortcomings of standard few-shot ICL in the presence of large and complex schemas. It combines two complementary techniques. First, \emph{schema pruning} dynamically selects a relevant sub-schema for each question, reducing contextual noise, shortening the prompt, and allowing schemas that exceed the LLM's context window to be handled. Second, \emph{AST-based re-ranking} uses a lightweight semantic parser (fewer than 500M parameters), adapted to process schemas in parallel, to produce an approximate SQL query for the question, and then re-ranks demonstrations retrieved by question embeddings according to the similarity of their normalized SQL abstract syntax trees to this approximation. This structural matching ensures that retrieved examples are both semantically relevant and syntactically aligned with the target query. Across monolingual and cross-lingual semantic parsing benchmarks, ASTReS improves over state-of-the-art retrieval baselines; ablations confirm that both pruning and AST re-ranking contribute to the gains, and the auxiliary parser imposes minimal computational overhead while reducing input length.

\section{Architectural Design Patterns for Text-to-SQL}
\label{sec:architecture}

This section reviews the architectural design patterns that recur across Text-to-SQL systems and offers practical implementation guidance. We group architectures into three broad families: (i) Transformer backbone adaptations that embed schema structure directly into attention and representation layers; (ii) attention and linking mechanisms that explicitly connect natural language tokens to schema elements; and (iii) multimodal fusion strategies that treat the schema, instance data, and text as distinct yet interoperable modalities. For each family, we summarize typical instantiations, trade-offs, and implementation details so that readers can evaluate alternatives for research or production use.

\paragraph{Transformer modifications for schema-aware representation.}
Vanilla Transformer encoders were designed for homogeneous token sequences and therefore struggle with long schema encodings and relational structure. Common modifications include relation-aware attention, which injects pairwise relation types (e.g., same-table, foreign-key, or primary-key) as attention biases~\cite{wang2019rat}; hierarchical encoders that process the question and the schema separately and fuse them only in later layers; and hybrid Graph-Transformer blocks that apply message passing over a schema graph before or during the Transformer layers. Relation-aware attention is typically implemented as an additive bias on the attention logits indexed by the relation type of each node pair; hierarchical encoders reduce the effective context length and allow schema embeddings to be precomputed and reused across queries; and graph-based encoders (GNNs or graph attention) produce compact node vectors that capture multi-hop schema topology. These modifications improve schema sensitivity at the cost of modest engineering and runtime overhead. Practitioners should choose the smallest modification that meets their correctness requirements; for many benchmarks, a relation bias combined with a hierarchical encoder is an effective compromise.

\paragraph{Attention mechanisms and explicit schema linking.}
Accurately mapping natural language mentions to table and column identifiers is central to Text-to-SQL. Architecturally, this is realized with targeted attention variants: cross-attention to explicit schema tokens, pointer/copy attention~\cite{vinyals2015pointer} to emit canonical column names, attention masking or focused attention to restrict language-to-schema interactions, and learned attention biases that encode relational priors. Practical systems combine a lightweight pruning stage (a classifier or retrieval step that selects a schema subset) with a fusion layer: the pruned schema tokens are injected as the keys and values of the decoder's cross-attention, while the question representation serves as the query. Pointer mechanisms are essential when exact-match fidelity is required, because they reduce tokenization-induced errors, whereas learned attention biases and masking schedules are useful for large schemas because they control the quadratic growth of attention. In short, pipelines often couple an inexpensive schema selector with an attention-rich fusion module to balance accuracy and cost.

\paragraph{Multimodal fusion, encoder--decoder choices, and a canonical pipeline.}
Treating the schema and instance values as separate modalities yields flexible designs: (a) \emph{inline fusion} concatenates schema tokens with the question and relies on a single encoder; (b) \emph{hierarchical fusion} encodes the question and schema separately and fuses them through cross-attention or a lightweight fusion network; and (c) \emph{graph fusion} uses node embeddings from a GNN as the keys and values of the decoder's attention. Decoder choices involve similar trade-offs: autoregressive sequence decoders are broadly compatible with LLM toolchains but can produce syntactically invalid SQL, whereas AST- or transition-based decoders enforce grammar constraints at generation time and therefore improve exact-match metrics. Algorithm~\ref{alg:canonical} gives a canonical implementation pattern that combines pruning, schema encoding, fusion, constrained decoding, and a single-step repair loop.

\begin{algorithm}[t]
\caption{Canonical Text-to-SQL pipeline (prune, encode, fuse, decode, validate)}
\label{alg:canonical}
\begin{algorithmic}[1]
\Require Natural language question $q$, full schema $S$, generation model $M$
\State $S' \gets \textsc{PruneSchema}(q, S)$ \Comment{selector or retriever}
\State $\mathbf{q} \gets \textsc{EncodeQuestion}(q)$
\State $\mathbf{S} \gets \textsc{EncodeSchema}(S')$ \Comment{Transformer or GNN}
\State $\mathbf{z} \gets \textsc{Fuse}(\mathbf{q}, \mathbf{S})$ \Comment{cross-attention or concatenation}
\State $\widehat{\mathrm{SQL}} \gets \textsc{DecodeConstrained}(M, \mathbf{z})$
\If{\textbf{not} \textsc{ValidateExec}($\widehat{\mathrm{SQL}}$)}
    \State $\widehat{\mathrm{SQL}} \gets \textsc{Repair}(M, \widehat{\mathrm{SQL}}, \mathbf{z})$
\EndIf
\State \Return $\widehat{\mathrm{SQL}}$
\end{algorithmic}
\end{algorithm}

\paragraph{Training recipes, hyperparameters, and complexity.}
For reproducibility and fair comparison, we recommend reporting at least the following: the base model and its parameter count; the optimizer (e.g., AdamW); the base learning rate and scheduler (e.g., linear warm-up followed by cosine or linear decay); the batch size and gradient-accumulation steps; the number of epochs or total gradient steps; the maximum token lengths for the question and the schema; and any PEFT or adapter settings (rank, bottleneck size). Commonly used settings include learning rates of $1\times10^{-5}$ to $3\times10^{-5}$ for full fine-tuning and around $1\times10^{-4}$ for LoRA adapters, an effective batch size of about 32, warm-up over 5--10\% of training steps, and a maximum schema budget of about 1,024 tokens after pruning; these should be treated as starting points rather than universal defaults. In terms of complexity, self-attention over a concatenated sequence of length $L$ requires $O(L^{2})$ time and memory, so when $L = T + S$ (text plus schema), the cost grows quadratically with schema size. Dual-encoder or hierarchical fusion reduces online cost by precomputing schema embeddings: a schema encoder followed by cross-attention fusion has an inference cost of approximately $O(T\cdot S_{\text{enc}} + T^{2})$, where $S_{\text{enc}}$ is the length of the compressed schema representation. Graph encoders add $O(E \cdot d)$ cost per message-passing round (for $E$ edges and embedding dimension $d$), which is linear in the number of schema relations. In deployment, common optimizations are schema pruning, cached schema encodings, PEFT adapters to reduce the memory footprint, and limited beam sizes to control decoding cost.

\paragraph{Recommendations and trade-offs.}
When correctness and exact-match fidelity are paramount (for example, in audit-critical analytics), hierarchical encoders, pointer/copy attention, and constrained or AST-based decoders are preferable, accepting higher implementation complexity in exchange for robustness. For low-latency, large-schema settings, dual-encoder designs with offline schema encoding and a lightweight schema pruner reduce the active context. Relation-aware attention and small GNN fusion modules provide robust structural priors and are worthwhile when schemas have complex foreign-key topologies. Finally, papers should report training and inference hyperparameters, an explicit complexity analysis, and an ablation that isolates the effect of schema-encoding choices; these are essential for reproducibility and for practitioners making informed engineering trade-offs.

\section{The Semantic Layer for Text-to-SQL}
\label{sec:semantic-layer}

\begin{figure}[t]
    \centering
    \includegraphics[width=0.62\linewidth]{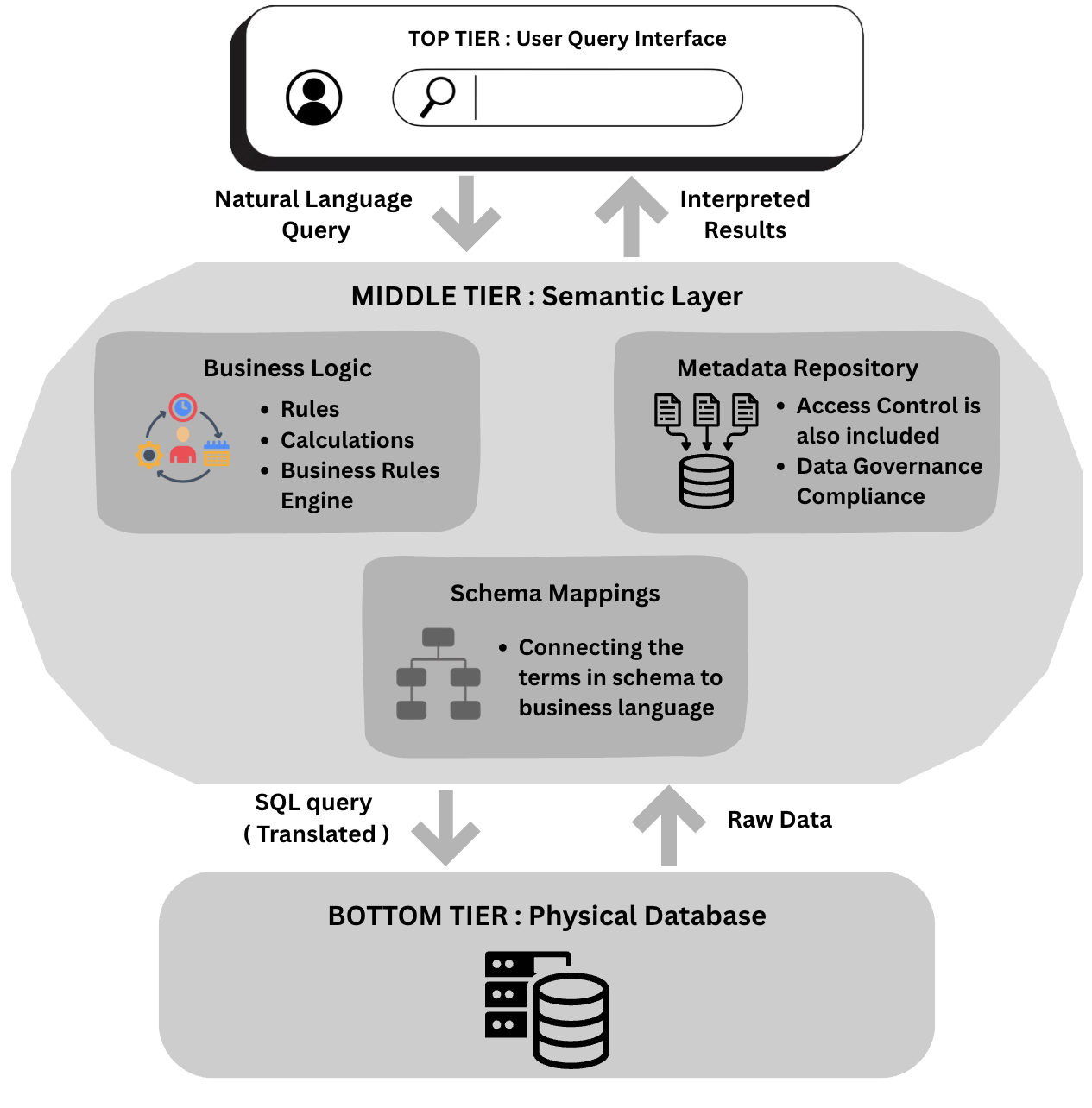}
    \caption{Three-tier semantic layer for Text-to-SQL systems. The middle
    tier mediates between the natural language interface and the physical
    database by encoding business logic, metadata, schema-to-business-term
    mappings, and access-control policies, thereby reducing schema
    ambiguity and enforcing governance constraints on generated SQL.}
    \label{fig:semantic_layer}
\end{figure}

A \emph{semantic layer} is an intermediate, formally specified mapping between high-level user concepts (business entities, metrics, hierarchies, synonyms, and constraints) and the physical database schema. In Text-to-SQL, the semantic layer serves as a domain-aware abstraction that (i) exposes a curated vocabulary to end users and models, (ii) encodes canonical definitions of metrics and dimensions, and (iii) constrains permissible joins and filters (Figure~\ref{fig:semantic_layer}). By decoupling intent representation from physical schema details, a semantic layer reduces ambiguity, improves the interpretability of generated queries, and provides a stable surface for governance and optimization. In short, it transforms the ill-posed task of ``translating arbitrary natural language into SQL'' into the better-posed task of ``mapping natural language to a governed logical model and then lowering that model to SQL.''

\subsection{Why a Semantic Layer Matters}
The following list enumerates common failure modes of na\"ive (non-semantic) Text-to-SQL systems and explains how a semantic layer addresses each one.

\begin{itemize}
  \item \textbf{Schema ambiguity and inconsistent naming.}
  \begin{description}[font=\normalfont\itshape, leftmargin=1.2em]
    \item[Failure:] Column and table identifiers vary widely across schemas and often do not reflect business meaning, so models misinterpret user terms (e.g., \texttt{cust\_id} vs.\ \texttt{client\_no}).
    \item[Mitigation:] A semantic layer provides canonical entity and attribute names, synonyms, and short human-readable descriptions mapped to the underlying columns, enabling robust resolution of user vocabulary to the correct schema elements.
  \end{description}

  \item \textbf{Implicit joins and relationship discovery.}
  \begin{description}[font=\normalfont\itshape, leftmargin=1.2em]
    \item[Failure:] LLMs frequently hallucinate joins or omit necessary join paths when relationships are not explicit in the prompt, producing incorrect or non-executable SQL.
    \item[Mitigation:] The semantic layer encodes the allowed relationships (predefined joins or view definitions), so translation proceeds over a constrained graph of entities rather than an unconstrained search over all tables.
  \end{description}

  \item \textbf{Domain-specific metrics and business logic.}
  \begin{description}[font=\normalfont\itshape, leftmargin=1.2em]
    \item[Failure:] Important constructs such as ``revenue,'' ``active users,'' or ``churn'' are computed inconsistently or misinterpreted by models that lack domain definitions.
    \item[Mitigation:] By defining metrics and their canonical SQL expressions in one place, the semantic layer enforces consistent computations and reduces semantic drift across queries and reports.
  \end{description}

  \item \textbf{Natural language variability and synonymy.}
  \begin{description}[font=\normalfont\itshape, leftmargin=1.2em]
    \item[Failure:] Users express the same intent with diverse phrasing (e.g., ``sales by region'' vs.\ ``revenue per territory''), which confuses direct NL$\rightarrow$SQL models.
    \item[Mitigation:] Registered synonyms, phrase mappings, and short glosses in the semantic layer normalize user utterances to a controlled vocabulary before translation to SQL.
  \end{description}

  \item \textbf{Schema evolution and fragility.}
  \begin{description}[font=\normalfont\itshape, leftmargin=1.2em]
    \item[Failure:] Direct mappings baked into prompts or model weights break when the physical schema changes (column renames, partitioning, denormalization).
    \item[Mitigation:] The semantic layer isolates higher-level definitions from their implementation, so changes to the underlying schema require only local updates to the layer rather than retraining or prompt rewriting.
  \end{description}

  \item \textbf{Governance, security, and access control.}
  \begin{description}[font=\normalfont\itshape, leftmargin=1.2em]
    \item[Failure:] Na\"ive systems expose raw schema details and allow unconstrained access that may violate privacy or compliance policies.
    \item[Mitigation:] A semantic layer can embed permission checks, row- and column-level masking rules, and approved metric definitions, enabling safe natural language access consistent with organizational policies.
  \end{description}

  \item \textbf{Search space and latency.}
  \begin{description}[font=\normalfont\itshape, leftmargin=1.2em]
    \item[Failure:] Translating directly to SQL over a large schema induces a large combinatorial search space, increasing inference cost and latency.
    \item[Mitigation:] By narrowing the search to a curated set of entities and precomputed aggregates (materialized metrics), the semantic layer reduces the candidate space and improves both accuracy and performance.
  \end{description}
\end{itemize}

\subsection{Semantic Layer via a Knowledge Graph: LinkedIn's Approach}

LinkedIn's production Text-to-SQL system~\cite{Chen2025EnterpriseSQL} grounds language models in a richly annotated \textbf{knowledge graph (KG)} that aggregates schema artifacts, documentation, query logs, example NL--SQL pairs, code snippets, and crowdsourced domain notes. The KG is implemented as a multi-indexed repository rather than a single RDF graph: \textbf{table} and \textbf{column} nodes store DDL-like metadata (types, foreign keys) together with human- and AI-written descriptions, sample cell values, certification and deprecation flags, and empirical usage statistics; \textbf{example-query} nodes link natural language samples to their execution traces; and \textbf{domain} nodes capture business jargon and canonical mappings. Edges encode explicit foreign keys, co-query and co-join statistics inferred from query plans and logs, and semantic-similarity links computed from embeddings. To scale to millions of tables, the system additionally computes usage-based soft clusters (200 clusters of 20 tables) via ICA over a user$\times$table access matrix; these clusters seed personalized retrieval and sharply reduce the irrelevant context passed to the LLMs.

Retrieval from the KG follows a conservative, multi-stage pipeline designed to balance token budget and semantic coverage. A \textbf{KG retriever} first seeds candidates using product-area clusters and explicit user mentions and then uses embedding-based indices (over tables, examples, and documentation) to fetch a broad candidate set. Lightweight LLM rankers (a table ranker and a column ranker) re-score and explain the candidates, and the \textbf{query-writer} agent receives only the top-ranked tables and columns and curated exemplars, from which it generates candidate SQL together with an explicit list of assumptions. An automated \textbf{researcher/fixer} validation loop checks the SQL in read-only mode (e.g., with Trino \texttt{EXPLAIN}/\texttt{VALIDATE}), detects schema hallucinations or runtime anomalies, and issues targeted repairs by fetching additional KG context. The production KG is refreshed at multiple cadences (near-real-time for user mappings and preferences, continuously for logs, and weekly for embeddings and clusters), so that indices remain fresh while update costs stay under control. The authors' ablations show that the KG and orchestration are essential in practice: moving from schema-only inputs to the full KG raises the rate of correct or near-correct responses from about 9\% to about 48\%, increases compile/validation success from about 66\% to about 96\%, and reduces the schema-hallucination rate from about 23\% to about 1\%. These results underline that a layered semantic index combined with execution-guided repair is critical for reliable, enterprise-grade Text-to-SQL.

\subsection{Semantic Layer via a Modeling Language: Wren AI's MDL}

Wren AI implements a production-oriented semantic layer through the \textbf{Wren Engine}~\cite{wrenai2024semanticlayer}, whose central abstraction is a declarative \textbf{Modeling Definition Language (MDL)}. MDL is a schema-level domain-specific language that encodes high-level business semantics independently of physical database layouts. It captures human-readable model and column names, rich descriptions, canonical business metrics, explicit inter-table relationships, and reusable calculation templates. Within a Text-to-SQL pipeline, MDL serves both as a semantic grounding mechanism for LLMs and as an operational layer that deterministically rewrites semantic queries into executable, dialect-specific SQL.

Methodologically, Wren's semantic layer follows a structured, engineering-driven workflow. First, MDL artifacts are authored or generated from existing enterprise metadata sources, including database schemas, documentation, and example queries; these artifacts define logical models, attributes, join relationships, and metric formulas. Second, the Wren Engine builds a runtime semantic index and query compiler. When a natural language question is issued, the engine resolves business terms to MDL entities, expands metric definitions into canonical SQL expressions, enforces approved join paths and access-control policies, and finally transpiles the logical query into backend-specific SQL (for example, for BigQuery, PostgreSQL, or Snowflake). This design reduces schema hallucinations, centralizes business logic, and isolates higher-level semantics from physical schema evolution.

\paragraph{Key capabilities of MDL-based semantic modeling.}
\begin{itemize}
  \item \textbf{Canonical metric definitions:} business metrics are defined once in MDL and expanded consistently across all generated queries, preventing semantic drift.
  \item \textbf{Explicit relationship modeling:} join paths, cardinalities, and conditions are declared explicitly, constraining SQL generation to valid and approved relationships.
  \item \textbf{Reusable macros and templates:} parameterized expressions provide concise representations of complex or repetitive calculations.
  \item \textbf{Deterministic SQL transpilation:} logical queries over MDL models are rewritten into dialect-specific SQL without relying on probabilistic generation.
  \item \textbf{Governance and access control:} row- and column-level permissions can be embedded directly in the semantic layer, enabling safe enterprise deployment.
  \item \textbf{Decoupling from physical schema changes:} updates to underlying tables or columns require only localized MDL changes, avoiding prompt rewrites or model retraining.
\end{itemize}

\paragraph{Implications for enterprise Text-to-SQL.}
By making semantics an explicit, version-controlled artifact, Wren's MDL-based approach turns Text-to-SQL from an unconstrained natural language translation problem into a governed semantic mapping task, improving robustness, interpretability, and maintainability in production. More broadly, it illustrates how a lightweight but expressive semantic layer can complement LLM reasoning while retaining the deterministic guarantees required by enterprise analytics systems.

\section{Responsible and Ethical AI Considerations}
\label{sec:responsible-ai}

As LLMs become central to Text-to-SQL pipelines, rigorous ethical and responsible AI practices become essential. These systems frequently access sensitive enterprise data and therefore raise concrete risks concerning privacy, data security, representational bias, and misuse. In this section, we discuss the principal dimensions of responsible AI in this domain: privacy-preserving data handling, access control and sandboxing, bias detection and mitigation, auditability and provenance, and operational governance of agentic behavior. Our discussion draws on recent audit studies and emerging governance practices for LLM-based agents and emphasizes practical measures that reduce risk while maintaining utility.

\subsection{Privacy and Data Protection}
Text-to-SQL systems often handle queries that may expose sensitive organizational or personal data. Effective data governance requires safeguards such as:
\begin{itemize}
    \item \textbf{Data anonymization and masking:} masking personally identifiable information (PII), such as social security or credit card numbers, in query outputs.
    \item \textbf{Differential privacy:} adding calibrated noise to aggregate outputs to prevent the reverse inference of sensitive records.
    \item \textbf{Least-privilege access:} restricting agents' database access through read-only roles, minimizing the potential damage from agent failures or misuse.
    \item \textbf{Role-based access control (RBAC):} ensuring that users can access only the data they are authorized to view, enforced through semantic abstractions (Section~\ref{sec:semantic-layer}).
\end{itemize}

\subsection{Transparency and Explainability}

Transparency and explainability are essential for user trust, systematic validation, and accountability, particularly in high-stakes domains such as healthcare, finance, and legal analytics. As LLMs take on increasingly autonomous roles through agentic pipelines, both end users and developers need to trace how modeling choices convert user intent into query operations. Clear explanations of schema-linking decisions, join paths, predicate selection, and aggregation logic help stakeholders verify correctness, identify failure modes, and make informed interventions.

A key challenge is that SQL generated by LLMs is often difficult for non-technical users to interpret: they may be unable to trace how a question was converted into particular tables, join paths, filters, or aggregations. To reduce this opacity, Text-to-SQL systems should expose intermediate representations and structured reasoning artifacts, such as schema-aligned semantic parse trees, clause-level rationales that justify selected predicates and aggregations, and controlled chain-of-thought annotations that document entity resolution and schema-linking decisions. These artifacts increase interpretability without forcing users to read raw SQL, making the system more transparent and more usable for non-technical stakeholders.

Practical implementations of explainability include:
\begin{itemize}
    \item \textbf{Query annotation:} annotating each SQL clause with a natural language explanation, e.g., highlighting that \texttt{WHERE year = 2023} was derived from the phrase ``in 2023'' in the user's question.
    \item \textbf{Step-by-step rationalization:} providing natural language justifications during SQL construction, such as ``To calculate average sales, the system applied \texttt{AVG(sales)} to the \texttt{sales} table filtered by region.''
    \item \textbf{Schema-alignment visualization:} presenting graphical or tabular views that show how specific phrases in the question were mapped to table and column names.
    \item \textbf{Conversational re-validation:} allowing users to ask follow-up questions or request clarifications, such as ``Why was the \texttt{orders} table used?'' or ``What does this filter condition mean?''
\end{itemize}

Explainability supports not only interpretability but also debugging, governance, and user education. In business intelligence platforms and decision-support dashboards, making explicit how filters, joins, and metrics are derived allows analysts to verify correctness and audit downstream decisions. As Text-to-SQL systems evolve toward more agentic and autonomous interaction, structured transparency and explanation mechanisms become a requirement for responsible and trustworthy deployment.

\subsection{Bias and Fairness}

Bias and fairness are critical concerns when Text-to-SQL systems support decision-making in domains such as finance, healthcare, education, and customer analytics. These systems inherit biases from their training data and from the broader socio-technical contexts in which they are developed. If not addressed, such biases can lead to SQL generation that skews analyses, reinforces disparities, or marginalizes under-represented groups.

Bias in Text-to-SQL may arise from several sources: (i) imbalanced datasets that over-represent certain domains, industries, or demographic attributes; (ii) skewed schema distributions that bias generation toward majority-table patterns; and (iii) annotation artifacts and spurious correlations in NL--SQL pairs. For instance, if training data disproportionately contain queries about ``high-income regions'' or ``top-selling products,'' a model may learn to prioritize these attributes even in neutral or unrelated questions, resulting in biased filtering or aggregation.

To promote fairness, Text-to-SQL systems should adopt the following strategies:
\begin{itemize}
    \item \textbf{Dataset diversity and curation:} ensuring that training corpora span varied domains, schema structures, query complexities, and linguistic patterns, which reduces overfitting to dominant schemas or phrasings.
    \item \textbf{Fairness-aware preprocessing:} re-weighting samples, up-sampling minority cases, and de-biasing schema mentions to balance the training signal across under-represented query forms and domains.
    \item \textbf{Auditing and evaluation:} conducting regular audits to detect systematic disparities in generated queries, for example by measuring whether a model disproportionately includes or excludes specific attributes (e.g., gender, region, or product line) across questions with similar intents.
    \item \textbf{Counterfactual testing:} generating near-identical questions with modified demographic or categorical terms (e.g., ``region = East'' vs.\ ``region = West'') to check whether the model treats them equitably.
\end{itemize}

In enterprise deployments, fairness violations can have cascading effects; for example, biased sales queries may lead to misallocated resources or unjustified strategic decisions. Text-to-SQL models should therefore be accompanied by governance frameworks that include fairness-aware training practices, transparent evaluation protocols, and the ability for human stakeholders to inspect, challenge, and override potentially biased outputs. These practices are essential not only for regulatory compliance but also for building equitable and trustworthy data-access systems.

\subsection{Human Oversight and Accountability}
While automation enhances scalability, human-in-the-loop mechanisms remain essential in high-stakes scenarios (see, e.g., the abstention-based design of RTS in Section~\ref{sec:agentic}). Organizations should:
\begin{itemize}
    \item allow users to review and override LLM-generated SQL in critical applications (e.g., finance and healthcare);
    \item maintain audit logs of model decisions and interactions; and
    \item define clear responsibilities across teams for the development, deployment, and monitoring of such systems.
\end{itemize}

\subsection{Security and Adversarial Robustness}
LLMs in Text-to-SQL pipelines are susceptible to several security threats:
\begin{itemize}
    \item \textbf{SQL injection:} natural language inputs can lead to the generation of malicious queries. Countermeasures include parameterized queries, read-only execution, and output validation.
    \item \textbf{Prompt injection:} attackers may manipulate prompts or embed instructions in database records that the model later reads. Input sanitization and post-generation query validation are essential.
    \item \textbf{Code-execution risks:} when agents can execute Python for post-processing, sandboxing must be enforced to prevent remote code execution.
    \item \textbf{Auditing frameworks:} tools such as CMPL (Conversational Manipulation for Privacy Leakage)~\cite{das2025cmpl} simulate multi-turn adversarial interactions to assess an agent's susceptibility to privacy leakage.
\end{itemize}

\subsection{Ethical Use and Societal Impact}
Deploying Text-to-SQL systems demands alignment with societal values and ethical norms. Developers should consider:
\begin{itemize}
    \item \textbf{Environmental impact:} optimizing models for energy efficiency to minimize their carbon footprint.
    \item \textbf{Inclusivity:} ensuring that interfaces and outputs are accessible to users of different abilities and linguistic backgrounds.
    \item \textbf{Avoiding misuse:} regularly reviewing deployment contexts to avoid applications that may harm the public interest (e.g., surveillance or manipulative analytics).
\end{itemize}

\subsection{Summary}
Responsible AI in Text-to-SQL is not a one-time compliance exercise but a continuous commitment to transparency, security, and societal well-being. As these systems are adopted across the enterprise and public sectors, integrating ethical practices into their full life cycle will be essential for long-term trust, scalability, and resilience.

\section{Best Practices for Text-to-SQL Systems}
\label{sec:best-practices}

To maximize the performance, reliability, and manageability of Text-to-SQL deployments, we distill the following best practices from recent industry guidelines (AWS~\cite{aws2024text2sql} and Google Cloud~\cite{google2024text2sql}), practitioner resources~\cite{turing2024text2sql}, and applied research.

\subsection{Caching and Monitoring}
\begin{itemize}
  \item \textbf{Result and prompt caching.} Cache parsed SQL statements together with their originating natural language prompts to avoid redundant model calls and SQL compilation. \emph{Rationale:} this reduces latency and compute cost for repeated or similar questions and keeps generated SQL consistent across sessions.
  \item \textbf{Comprehensive logging and metrics.} Instrument every stage (natural language understanding, prompt construction, SQL generation, execution time, and result cardinality) with logs and dashboards. \emph{Rationale:} this provides real-time visibility into system health, facilitates anomaly detection, and guides prompt or model refinements based on actual usage.
\end{itemize}

\subsection{Schema Management and Data Refresh}
\begin{itemize}
  \item \textbf{Materialized views for frequent queries.} Identify frequently accessed aggregates or joins and expose them as materialized views. \emph{Rationale:} this simplifies the SQL generation task and substantially improves runtime performance under heavy load.
  \item \textbf{Incremental view refresh.} Schedule batch or incremental refreshes of materialized views to keep derived data up to date. \emph{Rationale:} this balances data freshness against system overhead, ensuring timely answers without overloading the database.
  \item \textbf{Centralized data catalog.} Maintain a unified catalog of tables, views, and semantic metadata (column descriptions, business metrics) exposed to the LLM. \emph{Rationale:} this enables the model to select appropriate schema elements and supports retrieval-based prompt enrichment, improving query relevance and accuracy.
\end{itemize}

\subsection{Contextual Prompting and Disambiguation}
\begin{itemize}
  \item \textbf{Vector-based schema retrieval.} Use embeddings to dynamically retrieve the most relevant tables and columns for each question. \emph{Rationale:} this keeps prompts within context limits and focuses generation on the pertinent schema components.
  \item \textbf{Clarification dialogues.} When user intent is ambiguous (e.g., ``sales by region''), have the system ask a follow-up question such as ``Do you mean revenue or volume?'' before generating SQL. \emph{Rationale:} this reduces errors caused by underspecified questions and improves user satisfaction.
  \item \textbf{Entity resolution.} Resolve named entities and identifiers (e.g., customer names or product codes) against the database in advance to bind natural language mentions to actual records. \emph{Rationale:} this prevents failed or incorrect predicates arising from ambiguous or partial references.
\end{itemize}

\subsection{Prompt Engineering and Post-Processing}
\begin{itemize}
  \item \textbf{Dialect-specific few-shot examples.} Supply schema-aware in-context examples that reflect the organization's SQL dialect and naming conventions. \emph{Rationale:} this steers the model toward syntactically valid, idiomatic SQL for the target database engine.
  \item \textbf{Self-consistency and candidate voting.} Generate multiple SQL candidates and select the most frequent or highest-scoring one under a consistency criterion (ideally execution-result agreement). \emph{Rationale:} this mitigates the stochastic variability of LLM outputs and improves execution accuracy.
  \item \textbf{Dry-run validation.} Automatically parse and execute queries in a sandbox or read-only mode to catch syntax errors or empty results, then re-prompt the model with the error feedback. \emph{Rationale:} this provides a safety net that prevents invalid or dangerous queries from reaching production data.
\end{itemize}

\subsection{Synthetic Data Generation for Dialect- and Domain-Specific Settings}

High-quality labeled NL--SQL pairs are scarce for new dialects, regulated domains, and private enterprise databases. Three recent frameworks illustrate complementary ways to synthesize such data.

\paragraph{SQL-GEN.} SQL-GEN~\cite{pourreza2024sqlgen} is a dialect-aware synthetic data generation framework that addresses the gap between real-world SQL dialects (e.g., PostgreSQL and BigQuery) and the SQLite-centric focus of most prior Text-to-SQL research. Its core methodology combines tutorial-guided template induction, in which LLMs are conditioned on dialect documentation to learn dialect-specific syntax, functions, and constraints, with schema-grounded instantiation to generate large-scale, executable NL--SQL pairs. To consolidate multiple dialect-specialized models into a single deployable system, SQL-GEN proposes a Mixture-of-Experts (MoE) initialization strategy that merges attention parameters while preserving expert-specific feed-forward layers. A multi-stage filtering pipeline using SQLGlot validation, execution checks, and LLM-based quality assessment ensures syntactic and semantic correctness. SQL-GEN delivers consistent gains (up to about 20\% in execution accuracy) across dialect benchmarks and is strongly complementary to limited human-labeled data. Its novelty lies in explicitly modeling SQL dialects during synthetic generation and in a practical model-merging recipe for multi-dialect systems.

\paragraph{DSQG-Syn.} DSQG-Syn~\cite{duan2025dsqg} is a domain-constrained synthetic data generation framework designed to produce high-fidelity NL--SQL pairs for specialized and regulated databases where annotated data are scarce. Unlike query-first pipelines, it adopts a \emph{question-first} paradigm: domain-relevant questions are generated from schema keywords and a curated taxonomy of SQL reasoning types (e.g., aggregation, joins, and set operations), followed by schema-linking-driven SQL skeleton generation that decouples logical structure from schema instantiation. Multiple SQL candidates are instantiated per skeleton and paired with regenerated questions to enforce semantic consistency, after which an embedding-based semantic-optimization filter retains only the best-aligned pairs. Evaluated primarily by the execution accuracy of downstream parsers on OncoMX, CORDIS, and BIRD, with additional human and GPT-based relevance judgments, DSQG-Syn substantially improves over prior synthetic baselines and yields strong gains when combined with small real datasets.

\paragraph{SING-SQL.} SING-SQL~\cite{caferoglu2025singsql} targets the \emph{in-domain} setting typical of enterprise deployments, in which a model must specialize to a single, often large, database rather than generalize across domains. It is a fully automated two-stage framework that requires neither SQL logs nor manual annotations. The first stage partitions the target schema hierarchically into table-level and column-level sub-schemas to ensure broad coverage; the second follows a SQL-first, quality-aware generation pipeline that combines complexity-controlled SQL synthesis, LLM-as-a-judge validation, executability checks, automatic SQL repair, and reasoning-trace generation, with targeted re-generation for under-covered columns. The authors also release SingSQL-LM, a family of compact models (Qwen2.5-Coder 1.5B and 3B) fine-tuned with LoRA on the synthetic data, which achieve strong in-domain performance on a BIRD database and on the synthetic evaluation sets. SING-SQL thus offers a database-agnostic recipe for producing both training data and organization-specific evaluation sets.

\subsection{Debugging User SQL: SWE-SQL}

SWE-SQL~\cite{li2024swe} addresses the diagnosis and correction of erroneous SQL with respect to user intent. Whereas most LLM-based systems focus on generating SQL from natural language, this work targets debugging faulty user-written SQL.

\begin{itemize}
    \item \textbf{BIRD-CRITIC benchmark:} a debugging benchmark distilled from authentic user issues (e.g., StackOverflow bug-fix threads) and replayed in new environments, divided into a PostgreSQL-only subset (530 tasks) and a multi-dialect subset (570 tasks covering PostgreSQL, MySQL, SQL Server, and Oracle). Evaluation is based on functional correctness; even the strongest reasoning model evaluated, o3-mini, solves only 38.87\% and 33.33\% of the two subsets, respectively.
    \item \textbf{Training environment:} the \emph{Six-Gym} (SQL-fIX-Gym) environment creates training data automatically by reverse-engineering correct SQL into faulty variants, and \emph{f-Plan Boosting} guides models with high-level fix plans derived from the differences between faulty and correct SQL.
    \item \textbf{Bird-Fixer agent:} an open-source agent based on Qwen-2.5-Coder-14B, trained with Six-Gym and the SQL-Act scaffold, which generalizes well even to dialects not seen during training.
    \item \textbf{Results:} Bird-Fixer achieves 38.11\% success on the PostgreSQL subset and 29.65\% on the multi-dialect subset, surpassing leading proprietary models such as Claude-3.7-Sonnet and GPT-4.1.
\end{itemize}

\subsection{Content Balancing for RAG-Based Text-to-SQL}

Recent work~\cite{gurawa2025balancing} highlights the importance of the size and quality of retrieved content in retrieval-augmented Text-to-SQL pipelines. Excessive or poorly filtered context can confuse the model and lead to hallucinated SQL components, while insufficient context under-specifies the query. The following practices are recommended:

\begin{itemize}
    \item \textbf{Concise and relevant retrieval:} retrieved chunks should be semantically dense and strictly schema-relevant; redundant or verbose documentation increases token usage and degrades performance.
    \item \textbf{Retrieval filtering:} semantic-similarity thresholds and schema heuristics should be used to prune retrieved documents before they are included; in the reported experiments, moderately sized context (on the order of a few hundred tokens) offered the best trade-off between retrieval quality and hallucination.
    \item \textbf{Prompt confinement:} the model should be instructed to use only the schema elements present in the retrieved context, minimizing the chance of fabricating unseen structures.
    \item \textbf{Structured documentation:} schema descriptions should be well formatted and semantically grouped (e.g., mappings from tables to their purpose) to improve comprehension.
    \item \textbf{Evaluation strategies:} in addition to execution match, fuzzy and embedding-based SQL similarity, component-level matching (tables, columns, operations), and execution equivalence should be used to measure hallucination tendencies and correctness.
\end{itemize}

\noindent Together, these practices help mitigate hallucinations in RAG-based Text-to-SQL pipelines while preserving generation quality, especially in cross-domain or large-schema settings.

\subsection{Optimized Scheduling for Agentic Workflows: HEXGEN-TEXT2SQL}

Modern Text-to-SQL systems increasingly use agentic, multi-stage pipelines that sequence schema linking, candidate SQL generation, validation, and iterative correction. Each stage typically invokes LLM inference, creating cascaded dependencies that are sensitive to latency and to the availability of compute resources. Deploying such pipelines in enterprise settings therefore raises three interlinked challenges: (i) \emph{multi-stage dependency management}, because downstream modules depend on the timely and correct outputs of upstream stages; (ii) \emph{heterogeneous hardware orchestration}, because inference must be scheduled across GPU clusters with widely varying capabilities; and (iii) \emph{strict service-level objectives (SLOs)}, because per-query latency and throughput targets must be met in multi-tenant environments. Existing LLM-serving frameworks such as vLLM~\cite{kwon2023vllm} generally treat inference requests as independent tasks and provide no native support for coordinated scheduling across dependent stages or for optimizing placement under heterogeneous GPUs and hard SLOs.

To address these challenges, HEXGEN-TEXT2SQL~\cite{chen2025hexgen} introduces a hierarchical scheduling framework. At the global level, a dispatcher assigns each inference task $q_{i,j}$ to a GPU instance $m$ using a dispatch score that balances execution speed and load distribution. At the local level, each GPU instance maintains a dynamic priority queue that re-ranks tasks by \emph{urgency}, measured as deadline slack (the time remaining before an SLO violation). A lightweight simulator continuously analyzes execution traces to tune the hyperparameters of the scheduling policies, ensuring adaptability to workload shifts. Evaluated on multi-tenant workloads derived from BIRD, spanning domains such as finance and sports, on heterogeneous GPU clusters (A100, A6000, and L40), HEXGEN-TEXT2SQL reduces latency by up to $1.67\times$ (average $1.41\times$) and improves throughput by up to $1.75\times$ (average $1.65\times$) compared with vLLM, while maintaining consistently high SLO compliance. These results make workflow-aware scheduling a best practice for scalable, latency-sensitive deployment of agentic Text-to-SQL systems.

\section{Conclusion}
\label{sec:conclusion}

This survey has reviewed recent progress in Text-to-SQL along five complementary methodological paradigms: in-context learning; fine-tuning and knowledge-aware training; modular agentic and tool-augmented pipelines; retrieval-augmented generation; and reinforcement learning, including reasoning-based models. Together, these trends mark a significant shift from monolithic sequence-to-sequence parsers toward hybrid architectures that combine the generative capacity of large language models with structured schema awareness, targeted training signals, and lightweight specialist modules. In production systems, a well-designed semantic layer has emerged as a foundational component of this pipeline. The work reviewed here indicates that improvements in Text-to-SQL depend less on model scale alone and more on how models are grounded in schema and instance data, how retrieval and execution feedback are incorporated, how clean the training and evaluation data are, and how well evaluation metrics capture real-world failure modes.

Despite notable advances, several technical challenges remain. Reliable schema linking and robust value grounding continue to be the dominant failure modes: models frequently misalign natural language mentions with canonical table and column identifiers or assign literal values to the wrong fields. Execution fidelity is also unresolved, because syntactically valid SQL may still be semantically incorrect, and simple execution checks do not always surface deeper logical errors. Practical deployment exposes further scalability constraints: real-world schemas are large, heterogeneous, and constantly evolving, and systems must operate within context-window limits, latency targets, and monetary budgets that make many current methods difficult for organizations to adopt. Addressing these issues will require advances in grounding, verification, and cost-aware inference.

Evaluation practices must therefore remain multifaceted. Exact match is useful for assessing syntactic fidelity, while execution accuracy better reflects end-user correctness; robustness tests (paraphrase and synonym sets) and efficiency-aware metrics are needed to capture real-world utility, and expert-verified benchmarks are needed to ensure that reported gains are real. We advocate evaluation suites that report a balanced set of metrics (correctness, robustness, and cost/efficiency), together with difficulty-stratified analyses that make progress on hard, long-tail cases visible.

From an industry perspective, adoption depends critically on privacy, reliability, and operational cost. Solutions that limit data exposure, for example local or private fine-tuning, on-premise deployment, or small language models for sensitive stages, are likely to be adopted faster because of the data-security requirements of such systems. Production systems also require outputs that can be validated deterministically before execution, through schema-aware function calls, typed tool interfaces, or explicit provenance for generated queries. Finally, engineering trade-offs must be assessed against concrete product requirements: the balance between token cost and accuracy, and between the simplicity of a single-model solution and the maintenance burden of a multi-agent system, should be evaluated in terms of latency service-level agreements and safety constraints.

Several research directions deserve attention. First, improving cross-schema generalization and true zero-shot transfer to previously unseen databases would significantly broaden applicability and reduce per-database engineering. Second, integrating instance-level knowledge, such as cell values and representative rows, more tightly into model training and prompt-time grounding promises to reduce hallucinations and improve exact-match fidelity; agents that actively explore unknown schemas, rather than receiving them in full, are a promising step in this direction. Third, interactive and mixed-initiative interfaces, in which models propose candidate queries, ask clarifying questions, and abstain when uncertain while users provide lightweight corrections or confirmations, can increase reliability and user trust in high-stakes settings. Finally, evaluation metrics should evolve beyond binary correctness toward adaptive, usage-aware measures that capture cost, latency, explainability, and the practical usefulness of generated queries.

In summary, Text-to-SQL has evolved from a narrowly scoped research task into a full-stack systems problem that demands coordinated advances in model architectures, data engineering, tooling, and evaluation. Future progress will depend on principled grounding in both database schemas and instance data, efficient and auditable execution paths, and richer, cleaner robustness benchmarks, with a strong semantic layer as the foundation. When strong grounding, cost-aware inference, transparent validation, and deployment-oriented metrics are combined, Text-to-SQL systems can deliver reliable, scalable natural language access to structured data for both research and production use.

\bibliographystyle{plainnat}
\bibliography{references}

\end{document}